\documentclass[nopreprintline,12pt]{elsarticle}

\usepackage{amssymb}
\usepackage{amsmath}
\usepackage{cite}
\usepackage{amsmath,amssymb,amsfonts}
\usepackage{algorithmic}
\usepackage{textcomp}
\usepackage{amsmath,amsfonts}
\usepackage{algorithm}
\usepackage{tabularx}
\usepackage{array}
\usepackage[caption=false,font=normalsize,labelfont=sf,textfont=sf]{subfig}
\usepackage{textcomp}
\usepackage{stfloats}
\usepackage{url}
\usepackage{verbatim}
\usepackage{graphicx}
\usepackage[dvipsnames]{xcolor}
\usepackage{cite}
\usepackage{float}
\usepackage{booktabs}
\usepackage{placeins}
\usepackage{bm}
\usepackage{multirow}
\usepackage{bm}
\usepackage{amsmath}
\usepackage{threeparttable}
\usepackage{makecell}  

\journal{Knowledge-based Systems}

\begin{document}

\begin{frontmatter}



\title{Generative AI-Driven High-Fidelity Human Motion Simulation}

\author[label1,label3]{Hari Iyer}
\author[label2,label3]{Neel Macwan}
\author[label2]{Atharva Jitendra Hude}
\author[label1]{Heejin Jeong}
\author[label2]{Shenghan Guo}

\affiliation[label1]{
  organization={The Polytechnic School, Ira A. Fulton Schools of Engineering, Arizona State University},
  city={Mesa},
  postcode={85212},
  state={AZ},
  country={USA}
}

\affiliation[label2]{
  organization={School of Manufacturing Systems and Networks, Ira A. Fulton Schools of Engineering, Arizona State University},
  city={Mesa},
  postcode={85212},
  state={AZ},
  country={USA}
}

\affiliation[label3]{
  note={Hari Iyer and Neel Macwan contributed equally to this work.}
}




\begin{abstract}
Human motion simulation (HMS) supports cost-effective evaluation of worker behavior, safety, and productivity in industrial tasks. However, existing methods often suffer from low motion fidelity. This study introduces Generative-AI-Enabled HMS (G-AI-HMS), which integrates text-to-text and text-to-motion models to enhance simulation quality for physical tasks. G-AI-HMS tackles two key challenges: (1) translating task descriptions into motion-aware language using Large Language Models aligned with MotionGPT's training vocabulary, and (2) validating AI-enhanced motions against real human movements using computer vision. Posture estimation algorithms are applied to real-time videos to extract joint landmarks, and motion similarity metrics are used to compare them with AI-enhanced sequences. In a case study involving eight tasks, the AI-enhanced motions showed lower error than human created descriptions in most scenarios, performing better in six tasks based on spatial accuracy, four tasks based on alignment after pose normalization, and seven tasks based on overall temporal similarity. Statistical analysis showed that AI-enhanced prompts significantly (p $<$ 0.0001) reduced joint error and temporal misalignment while retaining comparable posture accuracy.
\end{abstract}




\end{frontmatter}



\begin{table}[H]
  \caption{Summary of mathematical notation used in this paper.}
  \vspace{0.1cm}
  \label{tab:nomenclature}
  \centering
  \renewcommand{\arraystretch}{1.2}
  \footnotesize
  \begin{tabularx}{\columnwidth}{|l|X|l|}
    \hline
    \textbf{Symbol} & \textbf{Definition} & \textbf{Units/Dimensions} \\
    \hline
    $g^{(i)}$                & Textual prompt for the $i$-th task                 & Text             \\
    \hline
    $\bm{D}^{(r)}_i$         & Real human motion video data for task $i$         & Video sequence   \\
    \hline
    $\bm{D}^{(s)}_i$         & G-AI-HMS simulated motion video for task $i$       & Video sequence  
    \\
    \hline
    $\bm{D}^{(b)}_i$         & MotionGPT-simulated motion video for task $i$       & Video sequence   \\
    \hline
    $\mathbf{M}^{\text{raw}}$& Raw motion sequence input                         & $T\times 22\times 3$ \\
    \hline
    $\boldsymbol{z}_e$       & Encoded latent motion vector                      & Vector           \\
    \hline
    $\boldsymbol{z}_q$       & Quantized latent code from VQ-VAE                 & Vector           \\
    \hline
    $\mathcal{E}$            & VQ-VAE codebook                                   & Codebook         \\
    \hline
    $\boldsymbol{e}_i$       & $i$-th codebook vector                            & Vector           \\
    \hline
    $E$, $D$                 & Encoder / Decoder functions                       & Function         \\
    \hline
    $\hat{\mathbf{M}}^{\text{recon}}$ 
                             & Reconstructed motion sequence                     & $T\times 22\times 3$ \\
    \hline
    $\mathcal{L}$            & Total VQ-VAE loss                                 & Scalar           \\
    \hline
    $\mathbf{P}_{t,j}$       & 3D position of joint $j$ at time $t$              & Vector           \\
    \hline
    $\tilde{\mathbf{P}}_{t,j}$ 
                             & Root-centered joint position                      & Vector           \\
    \hline
    $\hat{\mathbf{P}}_{t,j}$ & Normalized joint position                         & Vector           \\
    \hline
    $s_t$                    & Scale factor at time $t$                          & Scalar           \\
    \hline
    $\hat{y}_{t,j}$          & Flipped y-coordinate of joint $j$ at time $t$     & Scalar           \\
    \hline
    $\tilde{P}_{t,j,d}^{\text{filtered}}$ 
                             & Filtered value of joint $j$, axis $d$, at time $t$ & Scalar          \\
    \hline
    $f(t)$                   & Interpolation function                            & Function         \\
    \hline
    $x^{\text{resampled}}(t^*)$ 
                             & Resampled coordinate at time $t^*$                & Scalar           \\
    \hline
    $T$, $T^*$               & Frame count (raw, resampled)                      & Integer          \\
     \hline
    $T_r$                    & Number of frames in the ground truth motion sequence      & Integer \\
    \hline
    $T_b$                    & Number of frames in the human-observed motion sequence    & Integer \\
    \hline
    $T_s$                    & Number of frames in the model-generated motion sequence   & Integer \\
    \hline
    $\mathbf{A}, \mathbf{B}$ & Joint coordinate matrices for alignment           & Matrix           \\
    \hline
    $\bar{\mathbf{A}}, \bar{\mathbf{B}}$ 
                             & Mean-centered coordinate matrices                 & Vector           \\
    \hline
    $\mathbf{R}$             & Optimal rotation matrix (Procrustes)              & Matrix           \\
    \hline
    $\bar{\mathbf{M}}^{(g)}$ & AI-enhanced motion sequence                       & $T^*\times 22\times 3$ \\
    \hline
    $\bar{\mathbf{M}}^{(h)}$ & Human-prompt motion sequence                      & $T^*\times 22\times 3$ \\
    \hline
    $\bar{\mathbf{M}}^{(m)}$ & Ground truth (MediaPipe) motion sequence          & $T^*\times 22\times 3$ \\
    \hline
    MPJPE                    & Mean Per Joint Position Error                     & Distance         \\
    \hline
    PA-MPJPE                 & Procrustes-Aligned MPJPE                          & Distance         \\
    \hline
    DTW                      & Dynamic Time Warping score                        & Distance         \\
    \hline
  \end{tabularx}
\end{table}

\section{Introduction}
\label{sec:introduction}

Human motion simulation (HMS) is an essential technique for training, ergonomic evaluation, and safety analysis in manufacturing and industrial environments. HMS encompasses a variety of methods used to predict and visualize realistic human motions across different tasks and scenarios \citep{ref1, ref2, ref3}. As human workers continue to play a vital role in complex production systems, ensuring their physical safety and performance through accurate motion simulation becomes increasingly important. Accordingly, HMS is widely used in applications such as digital human modeling, industrial training, and scenario testing for risk mitigation.

Fidelity of simulated motion, defined as the degree to which simulated motion resembles real human behavior, is a central requirement for HMS effectiveness \citep{ref1, ref4}. Traditional simulation methods, including physics-based modeling, video-based tracking, and motion capture systems \citep{ref2, ref5}, are typically based on biomechanical rules and physical constraints. While these techniques provide a high degree of interpretability, they often require costly infrastructure, substantial time investment, and extensive expertise. They struggle with integrating contextual signals such as text instructions, environmental constraints, and object interaction, and they lack a scalable approach for validating simulated outputs against real-world behaviors \citep{ref1}.

Recent advances in artificial intelligence (AI) have introduced scalable and data-driven alternatives to traditional HMS \citep{ref61}. Generative AI models like Generative Adversarial Networks (GANs) \citep{ref6, ref67}, Diffusion Models \citep{ref8}, and Variational Autoencoders (VAEs) \citep{ref7} have shown significant potential for generating diverse and high-fidelity human motions from minimal inputs. These models are commonly referred to as text-to-motion (T2M) models because they synthesize full-body motions from natural language prompts \citep{ref16, ref42}. Among them, MotionGPT represents a recent state-of-the-art model that integrates a T5-based language model with a Vector Quantized Variational Autoencoder (VQ-VAE) motion tokenizer \citep{ref10, ref11}. It was trained on large datasets, such as HumanML3D, to generate temporally consistent and semantically accurate motion sequences from textual descriptions \citep{ref42}.

Despite these advancements, current evaluation metrics for T2M models are largely embedding-based and measure intra-model consistency rather than real-world fidelity. This presents a key limitation for deployment in safety-critical applications, where simulations must be verified against human reference data. Inconsistencies in textual inputs due to varied linguistic styles and out-of-distribution phrases can introduce ambiguity and reduce the quality of generated motions \citep{ref53}.

To address these challenges, this study introduces G-AI-HMS, a novel framework that uses LLMs such as ChatGPT to preprocess and standardize task prompts before passing them to MotionGPT \citep{ref38} (see Figure 1). This alignment ensures that the input language is consistent with the training distribution of the T2M model, thereby improving the fidelity of the generated motions. The study introduces a video-based evaluation framework that uses MediaPipe pose estimation to extract 3D landmarks from video recordings of human subjects \citep{ref17, ref51, ref54}. These landmarks are then spatially normalized, temporally resampled, and quantitatively compared to AI-enhanced outputs using Mean Per Joint Position Error (MPJPE), Procrustes-Aligned MPJPE (PA-MPJPE), and Dynamic Time Warping (DTW) metrics.

The resulting framework performs closed-loop validation of AI-enhanced motion simulations in controlled and real-world environments. This approach uses computer vision (CV) and joint-by-joint comparison to close the gap between AI-generated and human motion, making it easier to scale simulation tools for safety and training in industrial settings. Prior work outlined the general approach and presented preliminary findings of our research \citep{ref55}.

This study makes the following contributions:

1. It identifies limitations in prompt diversity and out-of-distribution generalization in existing T2M models, and addresses them through the use of AI-enhanced guidance.

2. It integrates LLM-based guidance generation and T2M-based motion synthesis, reducing the need for manual authoring of input prompts and improving motion quality.

3. It introduces a joint-wise evaluation framework to compare AI-enhanced motion sequences against human reference data using standardized metrics, namely MPJPE, PA-MPJPE, and DTW.

4. It demonstrates the adaptability of the proposed G-AI-HMS framework to multiple task domains, with the potential for deployment in real industrial environments using camera-based real-time sensing.

5. It highlights the importance of aligning prompt vocabulary with the motion model’s training distribution and proves that AI-enhanced prompts outperform human-written prompts in most scenarios.

The remainder of this paper is organized as follows: Section 2 reviews related work in Human Motion Simulation and generative AI, highlighting recent advancements in large language models and text-to-motion architectures. Section 3 describes the experimental design and data collection procedures, including the capture of human motion data and AI-enhanced sequences. Section 4 presents the G-AI-HMS methodology, detailing the integration of LLMs and MotionGPT, spatial-temporal alignment, and evaluation metrics. Section 5 outlines the data analysis pipeline, including motion normalization, resampling, and similarity metrics. Section 6 reports quantitative and qualitative results across eight distinct motion tasks. Section 7 discusses performance trends, anatomical fidelity, and prompt generation strategies. Finally, Section 8 concludes the paper and outlines directions for future work, including opportunities for hybrid prompting systems and multimodal fine-tuning. The symbols and terminology used throughout this paper are summarized in Table 1.

\begin{figure}[t!]
\centering
\includegraphics[width=\textwidth]{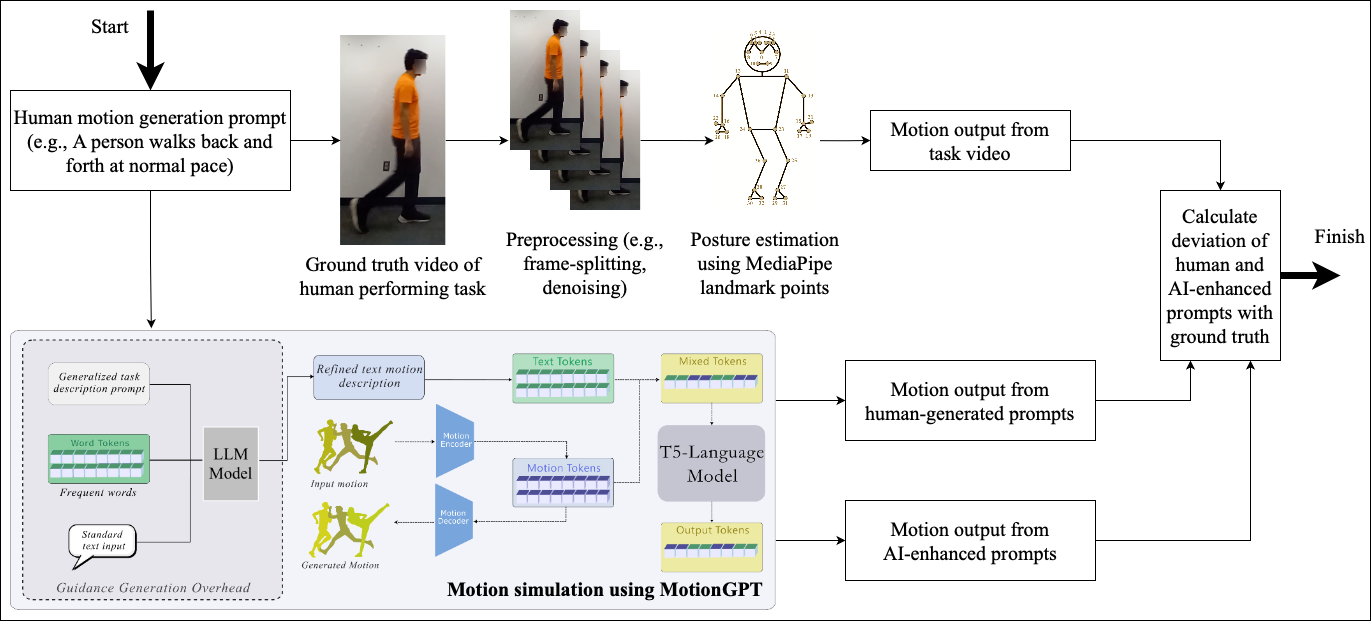}
\caption{Overview of the G-AI-HMS pipeline comparing motion from human-generated and AI-enhanced prompts against ground truth using MotionGPT and pose estimation \citep{ref11}.}
\end{figure}

\section{Literature Review}

\subsection{Human Motion Simulation Methods}

HMS has traditionally relied on a range of techniques that vary in precision, complexity, and data requirements \citep{ref2}. One foundational approach involves manually specifying the degrees of freedom for human joints across a sequence of keyframes. Motion is synthesized through interpolation algorithms that generate smooth transitions between frames. This method is precise and flexibile, but it takes a lot of time and requires expert knowledge to make it anatomically and biomechanically realistic.

Another widely used method is based on principles of biomechanics and physics. These simulations model human bodies as articulated systems with mass distributions, for motion that conforms to physical laws such as gravity, momentum, and joint torque. Although this physics-based modeling produces physically plausible movements, it can be limited in capturing expressive or non-rigid motions like dancing or gesturing.

Advances in computer vision have made it possible to estimate motion from video footage captured from one or more angles. In these video-based approaches, pose estimation algorithms are used to extract joint positions frame-by-frame. This provides a practical solution for motion acquisition in environments where motion capture equipment is unavailable. However, such techniques have limitations such as depth ambiguities, occlusions, and limited data reusability across different applications.

Data-driven simulation methods have become popular by relying on large collections of recorded human motion. These datasets, whether collected through manual annotation, physics simulation, or professional motion capture systems, are used to train models that generate new motions with high realism \citep{ref62}. While data-driven approaches benefit from the richness of pre-existing motion data, they often require expensive motion capture systems and extensive calibration, which may not be feasible for all users.

Hybrid methods that combine the strengths of multiple simulation techniques are also being actively explored. Prior studies, including those by Pan \& Zhang \citep{ref20}, Xiang et al. \citep{ref21}, and Fang \& Pollard \citep{ref22}, have shown that combining physics-based constraints with data-driven motion priors can generate realistic simulations with high degrees of freedom, suitable for ergonomic analysis and human–robot interaction.

These developments highlight an ongoing shift in HMS toward more intelligent, data-rich approaches. Specifically, the emergence of generative models marks a new phase in which high-fidelity human motion can be synthesized directly from compact inputs like text or voice, expanding the accessibility and scalability of motion simulation for a variety of real-world applications.

\subsection{AI for HMS}

AI provides algorithms and computational systems that help machines to perceive, interpret, and make decisions within dynamic environments to accomplish predefined tasks \citep{ref23}. With recent advancements in deep learning, data processing, and computational infrastructure, AI-based methods have become increasingly prominent in HMS \citep{ref1, ref55}. These approaches use large-scale motion datasets and high-capacity models to automate the synthesis, prediction, and evaluation of human movement.

A wide range of generative AI models have been applied to HMS. GANs \citep{ref6}, such as SA-GAN \citep{ref24}, Kinetic-GAN \citep{ref25}, and Text2Action \citep{ref26, ref27}, synthesize motion by training a generator-discriminator pair that refines outputs through adversarial feedback. Variational Autoencoders (VAEs) \citep{ref7}, used in models like Action2Motion \citep{ref26}, ACTOR \citep{ref28}, JL2P \citep{ref14}, TEMOS \citep{ref29}, and TEACH \citep{ref30}, learn low-dimensional probabilistic representations of motion and generate trajectories through sampling and decoding. More recently, diffusion-based models \citep{ref8}, including PhysDiff \citep{ref31}, MotionDiffuse \citep{ref32}, and RemoDiffuse \citep{ref33}, have demonstrated strong performance by iteratively denoising stochastic latent vectors to recover realistic motion sequences.

Generative models like GANs and diffusion frameworks provide a strong alternative to traditional rule-based or physics-based HMS methods, which often have trouble adapting to different situations or handling complex inputs. By learning a mapping from textual or visual inputs to motion trajectories, these models can produce high-fidelity simulations from natural language descriptions, audio cues, or 2D/3D pose estimates \citep{ref34}. Their neural-network-based design supports automatic evaluation of generated motion using quantitative metrics such as distance-based errors, angular deviations, or perceptual similarity \citep{ref1}.

However, a major limitation of these AI models is their dependence on large, labeled datasets for effective training, which can be labor-intensive and domain-specific \citep{ref35}. To mitigate this, the field is increasingly turning to pre-trained and open-access models that reduce the burden on end-users. These models, such as MotionGPT \citep{ref10, ref11} used in this study, generate zero-shot or few-shot motion from diverse inputs. Specifically, MotionGPT combines a VQ-VAE for motion tokenization with a T5-based language model to generate discrete motion sequences conditioned on natural language, for scalable and customizable HMS in ergonomic research, robotics, and animation.

\subsection{Generative Pre-trained Transformers (GPT) and MotionGPT}

Generative Pre-trained Transformers (GPT) represents a major advancement in natural language processing, for the generation of coherent and contextually appropriate responses from textual input \citep{ref40, ref39, ref53}. Based on the transformer architecture and implemented in its fourth iteration as GPT-4, it is trained on a large corpus of internet text \citep{ref36, ref37}. ChatGPT uses multi-head attention and positional encoding to model long-range dependencies in language, which makes it applicable for a wide range of tasks including prompt generation, summarization, and dialogue \citep{ref47}.

MotionGPT is a model used to simulate human motion generation from natural language \citep{ref10, ref11}. It uses a multi-stage architecture that first tokenizes motion using a VQ-VAE and then uses a text-to-text transformer (based on T5) to map linguistic input to motion token sequences. Unlike conventional motion capture methods that rely on hardware and calibration \citep{ref15, ref48}, MotionGPT synthesizes motion from text alone, eliminating the need for physical sensors or manual annotations. Its training pipeline includes three phases: (1) pretraining the motion tokenizer (VQ-VAE) on continuous 3D motion data to produce discrete motion tokens, (2) training the T5-based language model on aligned motion-text pairs, and (3) fine-tuning on prompt-instruction pairs to enhance instruction-following performance.

The VQ-VAE module encodes motion sequences into latent vectors, quantizes them via a learned codebook, and decodes them to reconstruct motion. This aids MotionGPT to operate in a discrete token space while maintaining fidelity to high-dimensional motion trajectories. Unlike recurrent neural networks \citep{ref49}, which model sequences in a stepwise manner, transformers model all time steps simultaneously, making them more efficient and better suited for capturing global dependencies in motion dynamics.

By integrating AI-enhanced verbal guidance with MotionGPT’s motion decoder, the system translates instructions such as ``a person walks and sits on a chair'' into smooth, realistic motion sequences. This close link between language and motion lets the system create flexible, text-based simulations that make sense in context, stay consistent over time, and can be checked against real human motion data.

\subsection{HMS with Generative AI}

HMS has traditionally relied on a range of techniques, including manual keyframe animation, physics-based modeling, video-based motion capture, and data-driven methods built upon pre-recorded datasets. While each approach has distinct benefits, they also present limitations related to realism, scalability, and cost. The advent of Generative AI models, such as GANs \citep{ref6, ref24, ref25}, Variational Autoencoders (VAEs) \citep{ref7}, and diffusion-based methods \citep{ref8}, help generate human motion content from minimal or abstract inputs.

Recent advancements have extended these capabilities further through the integration of LLMs such as GPT \citep{ref39, ref40} with text-to-motion architectures like MotionGPT \citep{ref10, ref11}. These systems translate complex, natural language descriptions into structured textual representations aligned with the motion vocabulary of pretrained models. MotionGPT uses a VQ-VAE-based motion tokenizer trained on the HumanML3D dataset to convert motion trajectories into discrete latent codes. These codes, embedded within a shared motion-language space, help the model to generate fluid and contextually accurate motion sequences that represent the semantics of the input descriptions.

The combination of LLM-driven prompt enhancement with motion-aware generative models could improve the quality and relevance of synthesized motion. AI-enhanced motions can be validated against real human motion data using CV techniques, such as landmark extraction via MediaPipe \citep{ref41}. This supports quantitative assessment of fidelity using joint-wise similarity metrics (e.g., MPJPE, PA-MPJPE, DTW), supporting evaluation of how closely the simulated motion aligns with real-world human performance.

These features show that generative AI could be a strong tool for creating realistic human motion at scale, useful in areas like workplace ergonomics, robotics, training, and virtual human modeling \citep{ref59, ref57}.

\section{Experiment Design and Data Collection}

This study adopts a data-driven approach based on three primary data types: (1) textual task guidance, denoted as $g$; (2) videos of human participants performing tasks, denoted as $\bm{D}_{(r)}$; and (3) AI-enhanced simulation videos, denoted as $\bm{D}_{(s)}$. The experimental protocol was designed to produce and align these datasets in support of subsequent evaluation tasks, including AI motion fidelity validation.

Each experimental trial involved a human subject performing a predefined task within a controlled lab environment. Prior to each performance, the task was specified verbally and transcribed as textual input, providing guidance data $g^{(i)}$ corresponding to the $i$-th task. Modeling such task instructions as natural language inputs aligns with approaches that emphasize multi-level semantic and emotional understanding to enhance machine interpretation of human intentions \citep{ref68}. As the subject executed the task, the motion was captured using a tri-camera setup (left, center, and right perspectives), generating the video $\bm{D}^{(r)}_i$.

To create a complete evaluation tuple, the same textual prompt $g^{(i)}$ was provided to the G-AI-HMS system, which used GPT-based guidance and MotionGPT for motion generation. This process produced a corresponding AI-simulated motion $\bm{D}^{(s)}_i$. This approach aligns with recent advances in socio-physically grounded motion modeling, such as SPU-BERT, which integrates scene context, social interaction, and goal-directed behavior for efficient multi-trajectory prediction \citep{ref66}.

For each task \( i \in \{1, 2, \dots, 8\} \), we generated three motion sequences: 
(i) a reference motion video \( \bm{D}_i^{(r)} \), extracted from human task demonstrations using MediaPipe \citep{ref51}, 
(ii) a benchmark motion video \( \bm{D}_i^{(b)} \), generated using the original human-written prompt via MotionGPT \citep{ref11}, and 
(iii) a G-AI-HMS motion video \( \bm{D}_i^{(s)} \), generated using the LLM-enhanced version of the same prompt. 
These triplets formed the basis for comparative evaluation across tasks.

\section{Methodology}

The novelty of this study lies in two key components: (1) the integration of text-to-text (T2T) and T2M generative AI models for task-level motion simulation, and (2) a quantitative evaluation of AI-enhanced motions against reference human data. This section outlines the methods used to generate and validate synthetic human motions, focusing on how ChatGPT was used to reconstruct text prompts and how MotionGPT was used to simulate corresponding motion sequences. We also describe the design of the human subject experiments used for ground truth collection and AI model validation.

\subsection{Human Subject Experiments}

\begin{table}
  \centering
  \caption{Description of the eight predefined tasks used for motion generation and evaluation, designed to cover a range of locomotion, manipulation, and posture transition behaviors.}
  \vspace{0.1cm}
  \begin{tabularx}{\columnwidth}{|c|X|}
    \hline
    \textbf{No.} & \textbf{Task Description} \\
    \hline
    1 & ``A person walks back and forth at normal pace.'' \\ \hline
    2 & ``A person throws a large object with their right hand.'' \\ \hline
    3 & ``A person walks and then sits on a stationary chair.'' \\ \hline
    4 & ``A person applies paint on a surface with a brush.'' \\ \hline
    5 & ``A person walks while carrying a box on top of their head.'' \\ \hline
    6 & ``A worker throws a small object with their right hand.'' \\ \hline
    7 & ``A person walks and then they sit on a swivel chair.'' \\ \hline
    8 & ``A person paints a wall using a brush.'' \\
    \hline
  \end{tabularx}
  \label{tab:task_descriptions}
\end{table}

Eight predefined tasks were developed to capture a representative range of full-body human motions relevant to occupational and everyday activities (see Table 2). The tasks were designed to span locomotion, manual object manipulation, posture transitions, and repetitive upper-body actions, for evaluation of both dynamic and static ergonomic risk factors.

Each task was performed by a human subject in a controlled indoor setting and recorded using a three-camera setup placed at left, center, and right viewpoints for comprehensive spatial coverage. The recording environment featured consistent lighting and a uniform background to support high-quality pose estimation. Videos were captured at 30 frames per second in high-definition resolution, ensuring temporal fidelity and spatial clarity for motion analysis.

Tasks were selected to be representative of different biomechanical profiles. Task 1 simulated load-carrying while walking, relevant to material handling; Task 2 and Task 6 examined variations in overhead throwing motion; Task 3 and Task 7 involved walk-to-sit transitions using different types of chairs; and Tasks 4 and 8 captured paint application motions against horizontal and vertical surfaces, respectively, for comparisons across workspace orientations. Task 5 required the subject to walk while balancing a box on the head, introducing constraints related to upper-body stability and head posture. This set of activities supported the generation and evaluation of diverse motion patterns across occupationally relevant scenarios. This research received approval from the Institutional Review Board at Arizona State University (STUDY00016442). Informed consent in written form was obtained from the participant.

\subsection{Real-time Sensing for Human Subjects}

Real-time sensing was conducted to obtain high-quality motion data from human subjects performing predefined tasks listed in Table 2. Each task was recorded using a tri-camera setup comprising left, center, and right viewpoints, arranged to provide multi-angle coverage of the subject's movements. This multi-perspective configuration was designed to aid the subsequent 2D-to-3D pose estimation by reducing occlusions and capturing full-body articulation.

All recordings were carried out in a controlled indoor environment to ensure consistency in lighting, background, and spatial reference. Tasks were performed against a uniform background to facilitate keypoint extraction and to minimize visual noise. Each video sequence was captured at 30 frames per second in high-definition resolution to preserve fine-grained motion characteristics necessary for accurate downstream analysis.

The selected tasks represented a diverse range of whole-body actions, including locomotion (e.g., walking), object interaction (e.g., throwing, lifting), seated movements (e.g., sitting), and manual activities (e.g., painting). These tasks were chosen to reflect varied kinematic and ergonomic profiles, for benchmarking of AI-enhanced motion against ground-truth human behavior.

\begin{table*}
\centering
\caption{Examples of original task prompts, ground truth input images, AI-enhanced prompts, and simulated motion outputs.}
\vspace{0.3em}
\scriptsize
\renewcommand{\arraystretch}{1.1}
\setlength{\tabcolsep}{4pt}
\begin{tabularx}{\textwidth}{
|>{\centering\arraybackslash}m{0.5cm}
|>{\centering\arraybackslash}m{3.5cm}
|>{\centering\arraybackslash}m{2.2cm}
|>{\centering\arraybackslash}m{3.8cm}
|>{\centering\arraybackslash}m{2.2cm}|
}
\hline
\textbf{No.} & \textbf{Human-generated Prompts} & \textbf{Ground Truth} & \textbf{AI-enhanced Prompts} & \textbf{MotionGPT Simulations} \\
\hline
1 
& ``A person walks back and forth at normal pace.''
& \vspace{0.1cm} \includegraphics[width=1.5cm, height=2.0cm]{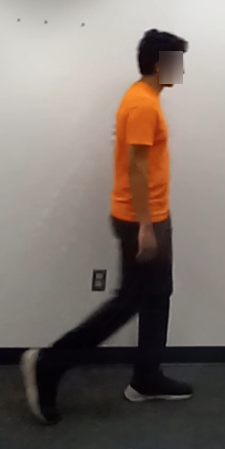}
& ``A person walks back and forth at a normal pace for several seconds on a flat surface.''
& \vspace{0.1cm} \includegraphics[width=1.5cm, height=2.0cm]{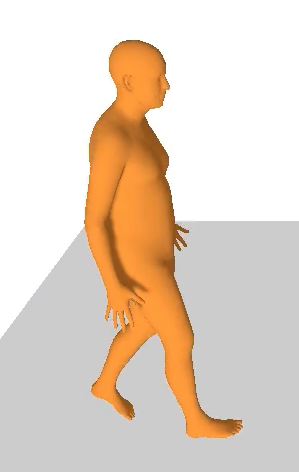} \\
\hline
2 
& ``A person throws a large object with their right hand.''
& \vspace{0.1cm} \includegraphics[width=1.5cm, height=2.0cm]{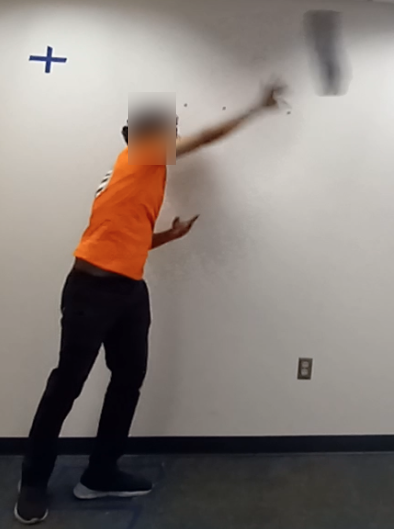}
& ``A person picks up a large object from the ground and throws it forward using their right hand.''
& \vspace{0.1cm} \includegraphics[width=1.5cm, height=2.0cm]{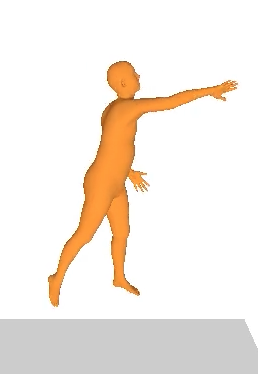} \\
\hline

3 
& ``A person walks and then sits on a stationary chair.''
& \vspace{0.1cm} \includegraphics[width=1.5cm, height=2.0cm]{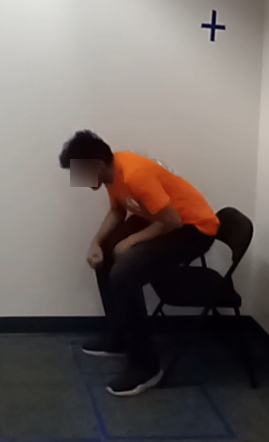}
& ``A person walks toward a stationary chair, pauses briefly, and then sits down in a controlled manner.''
& \vspace{0.1cm} \includegraphics[width=1.5cm, height=2.0cm]{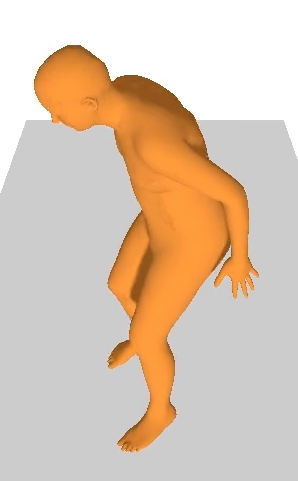} \\
\hline

4 
& ``A person is applying paint on a surface with a brush.''
& \vspace{0.1cm} \includegraphics[width=1.5cm, height=2.0cm]{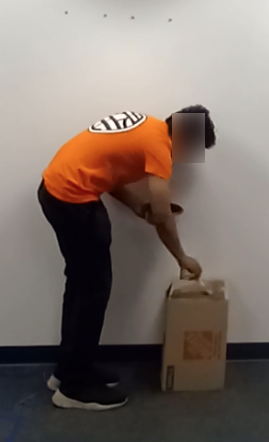}
& ``A person dips a brush into a paint bucket and applies paint on a large horizontal surface using repetitive strokes.''
& \vspace{0.1cm} \includegraphics[width=1.5cm, height=2.0cm]{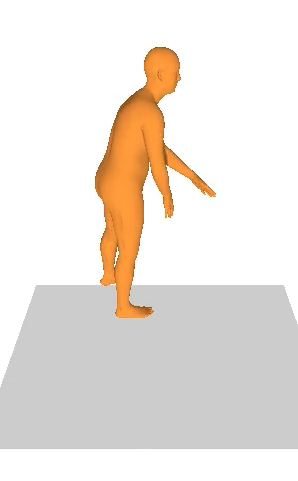} \\
\hline

5 
& ``A person is walking while carrying a box on top of their head.''
& \vspace{0.1cm} \includegraphics[width=1.5cm, height=2.0cm]{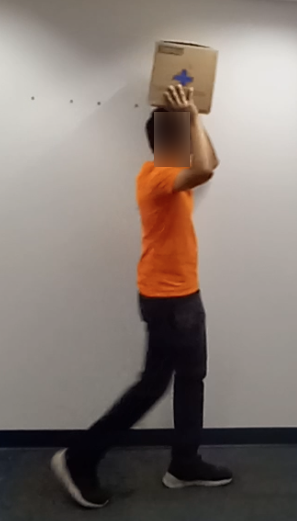}
& ``A person balances a medium-sized box on top of their head and walks slowly and steadily across a room.''
& \vspace{0.1cm} \includegraphics[width=1.5cm, height=2.0cm]{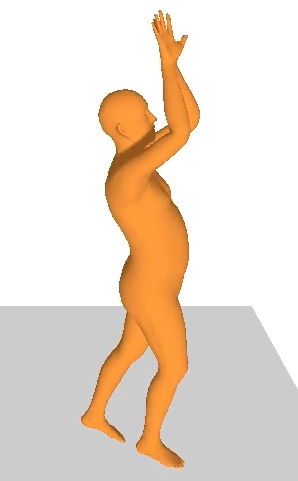} \\
\hline

6 
& ``A worker is throwing a small object with their right hand.''
& \vspace{0.1cm} \includegraphics[width=1.5cm, height=2.0cm]{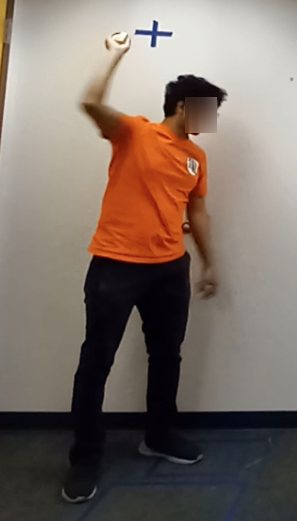}
& ``A worker swiftly picks up a small object from a table and throws it overhand using their right hand.''
& \vspace{0.1cm} \includegraphics[width=1.5cm, height=2.0cm]{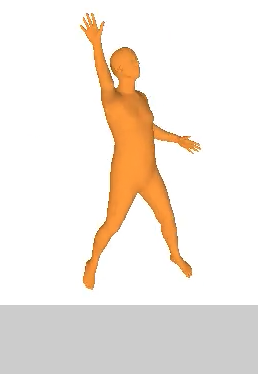} \\
\hline

7 
& ``A person is walking and then they sit on a swivel chair.''
& \vspace{0.1cm} \includegraphics[width=1.5cm, height=2.0cm]{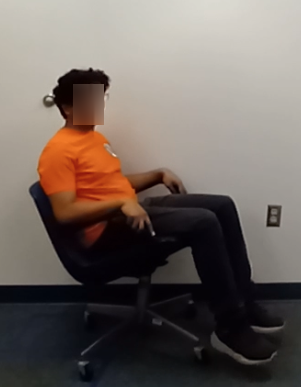}
& ``A person walks toward a swivel chair, turns around, and gently lowers themselves into a seated position while the chair slightly rotates.''
& \vspace{0.1cm} \includegraphics[width=1.5cm, height=2.0cm]{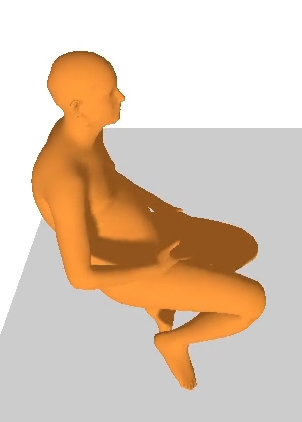} \\
\hline

8 
& ``A person is painting a wall using a brush.''
& \vspace{0.1cm} \includegraphics[width=1.5cm, height=2.0cm]{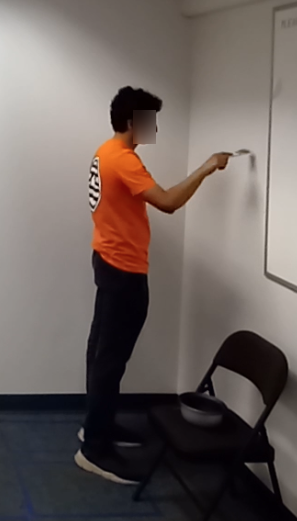}
& ``A person uses a handheld brush to paint a vertical wall using long upward and downward strokes.''
& \vspace{0.1cm} \includegraphics[width=1.5cm, height=2.0cm]{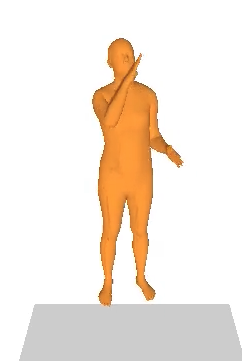} \\
\hline

\end{tabularx}
\label{tab:ai_simulation_examples}
\end{table*}

\subsection{Method Overview}

AI systems, such as MotionGPT \citep{ref11}, have been applied to simulate human motion from text prompts. However, generating accurate high-fidelity simulations remains challenging due to the disconnect between natural language inputs and the constrained vocabulary used during the training of text-to-motion models \citep{ref4}. This mismatch often leads to errors when AI models attempt to interpret vague or stylistically diverse task descriptions, resulting in semantically inaccurate or structurally flawed motion sequences.

To address these limitations, this study introduces a two-stage pipeline called \textit{Generative-AI-Enabled Human Motion Simulation (G-AI-HMS)}. In the first stage, a LLM, specifically GPT, is used to transform free-form task descriptions into structured guidance prompts. These prompts are composed using the most frequent action keywords from the HumanML3D dataset \citep{ref42}, which also serves as the training corpus for MotionGPT \citep{ref11}. By aligning the lexical structure of input prompts with the model’s training distribution, this approach minimizes out-of-vocabulary discrepancies and increases the likelihood of generating contextually appropriate motions. In the second stage, MotionGPT uses these refined prompts to produce corresponding 3D motion simulations in the form of temporally sequenced joint trajectories.

To benchmark our method against baselines, we used MotionGPT, a state-of-the-art text-to-motion model that achieves top performance across HumanML3D and KIT-ML datasets \citep{ref11}. We used ChatGPT (GPT-4) for prompt generation, as recent comparative studies identify GPT-4 as the most capable LLM for grounded task planning and instruction understanding in physical domains \citep{ref65}. These models are the leading tools for motion generation and language interpretation, and are used to benchmark AI-driven human motion simulation.

To evaluate the fidelity of G-AI-HMS, we conducted quantitative comparisons between the AI-enhanced sequences ($\bm{D}_i^{(s)}$) and reference ground truth sequences ($\bm{D}_i^{(r)}$) extracted from real human videos (see Table 3). The comparison used three established metrics that evaluate motion similarity. MPJPE was used to measure the average Euclidean distance between generated and reference joint positions across all frames. PA-MPJPE accounted for rigid-body transformations by aligning poses before computing positional errors. DTW quantified the temporal alignment cost between corresponding joint trajectories. These metrics were applied both at the full-body level and on a per-joint basis to ensure localized and global evaluations.

In addition to comparing $\bm{D}_i^{(s)}$ and $\bm{D}_i^{(r)}$, we also compared both against baseline sequences $\bm{D}_i^{(b)}$ that were generated from human-generated prompts. This triangulation assessed the added value of GPT-based guidance and the overall performance of the G-AI-HMS framework relative to traditional input strategies. Detailed descriptions of the computational process for each metric are provided in the Data Analysis section.

To ensure comparability across all sources of motion data, we included two additional stages prior to metric computation. All motion sequences underwent spatial normalization and smoothing. Trajectories were root-centered using the head joint, rescaled based on subject height, and denoised using a combination of median filtering and Butterworth low-pass filtering. These steps corrected for positional drift and anatomical inconsistency, improving the biomechanical fidelity of both real and generated motions. The full preprocessing procedure is detailed in Section 5.1.

All motion sequences were temporally aligned to a common length for frame-wise comparisons. We determined the minimum frame count \( T^* \) across each triplet \( \{\bm{D}_i^{(r)}, \bm{D}_i^{(b)}, \bm{D}_i^{(s)}\} \) and resampled all sequences via linear interpolation to this length. The temporal alignment strategy is described in Section 5.2.

\subsection{Guidance Generation with LLM}

LLMs are well-suited for knowledge-intensive tasks due to their ability to generalize across diverse contexts. Fine-tuned variants of these models, trained further on domain-specific corpora, have demonstrated strong performance in specialized domains such as finance, medicine, and law. However, in this study, we did not rely on fine-tuning. Instead, we explored the use of general-purpose LLMs for guiding motion synthesis in the context of factory-oriented human actions (e.g., lifting, picking, placing, carrying) \citep{ref58}.

To improve the realism and relevance of generated motions, we used GPT to produce semantic guidance for MotionGPT. This was done by integrating frequent action-related keywords derived from the HumanML3D dataset \citep{ref42}, which also served as the foundational training corpus for MotionGPT \citep{ref11}. These instructions generated by ChatGPT (GPT-4) were then used as input to the G-AI-HMS pipeline for motion generation.

\subsection{Motion Synthesis Using MotionGPT}

For AI-enhanced motion data, we used MotionGPT \citep{ref11}, a state-of-the-art text-to-motion generation model. MotionGPT uses a VQ-VAE as its motion tokenizer and a T5-based language model to convert natural language prompts into motion sequences. The motion generation pipeline begins by transforming the motion sequence $\mathbf{M}^{\text{raw}}$ into a latent representation $\boldsymbol{z}_e$ using the VQ-VAE encoder $E$, as shown in Equation (1).
\begin{equation}
\boldsymbol{z}_e = E(\mathbf{M}^{\text{raw}})
\end{equation}
Next, the latent vector $\boldsymbol{z}_e$ is quantized to its nearest codebook vector $\boldsymbol{z}_q$ by minimizing the Euclidean distance over the codebook $\mathcal{E} = \{ \boldsymbol{e}_1, \dots, \boldsymbol{e}_K \}$, as shown in Equation (2).
\begin{equation}
\boldsymbol{z}_q = \arg\min_{\boldsymbol{e}_i \in \mathcal{E}} \| \boldsymbol{z}_e - \boldsymbol{e}_i \|_2
\end{equation}
The quantized representation $\boldsymbol{z}_q$ is then fed into the decoder $D$ to generate the reconstructed motion sequence $\hat{\mathbf{M}}^{\text{recon}}$, as shown in Equation (3). During inference, the process is reversed: the input prompt \( g^{(i)} \) is tokenized using a T5-based language model and mapped to a sequence of latent motion tokens using the learned text-to-motion transformer. These motion tokens correspond to entries in the codebook \( \mathcal{E} \), producing a sequence of quantized latent vectors \( \boldsymbol{z}_q \). This sequence is then decoded by the VQ-VAE decoder \( D \), yielding the raw motion sequence \( \mathbf{M}^{\text{raw}} \), which captures joint trajectories over time. Thus, \( \mathbf{M}^{\text{raw}} \) represents the output of the MotionGPT pipeline when prompted with \( g^{(i)} \).

\begin{equation}
\hat{\mathbf{M}}^{\text{recon}} = D(\boldsymbol{z}_q)
\end{equation}

It is important to note that \( \hat{\mathbf{M}}^{\text{recon}} \) is used only during VQ-VAE training to reconstruct an encoded motion sequence and minimize reconstruction loss. \( \bm{D}_i^{(b)} \), the benchmark motion output of MotionGPT, is generated during inference from the text prompt \( g^{(i)} \), without involving an encoder. The prompt is first tokenized via a pretrained language model and passed through a transformer that predicts a sequence of codebook indices, which are then decoded via the VQ-VAE decoder. \( \bm{D}_i^{(b)} \) and \( \hat{\mathbf{M}}^{\text{recon}} \) differ in both origin and use: one is inference-time output, the other a training-time reconstruction.

The VQ-VAE is trained using the total loss $\mathcal{L}$, which comprises a reconstruction loss, a codebook loss, and a commitment loss weighted by $\beta$, as defined in Equation (4) \citep{ref43}.
\begin{equation}
\mathcal{L} = \| \mathbf{M}^{\text{raw}} - \hat{\mathbf{M}}^{\text{recon}} \|^2 + \| \text{sg}[\boldsymbol{z}_e] - \boldsymbol{e} \|^2 + \beta \| \boldsymbol{z}_e - \text{sg}[\boldsymbol{e}] \|^2
\end{equation}
Here, $\text{sg}[\cdot]$ denotes the stop-gradient operator, and $\beta$ is a scalar hyperparameter that controls the strength of the commitment loss. This VQ-VAE-based tokenizer helps MotionGPT to effectively learn a shared representation between language and motion tokens, with which it generates semantically consistent and biomechanically plausible sequences from instructional prompts. The generated motion from MotionGPT, denoted $\mathbf{M}^{(g)}$, was directly used for downstream comparison against MediaPipe and human-prompt-derived motions.

While inferencing, MotionGPT generates motion sequences directly from enhanced prompts \( g^{(i)} \) without needing input motion data. Each prompt is tokenized using a T5-based language model and decoded using the trained VQ-VAE to generate the output sequence \( \bm{D}_i^{(s)} \). The equations above describe the training-time behavior of the VQ-VAE component, using which the model learns a shared representation between motion and language. The AI-enhanced sequence \( \bm{D}_i^{(s)} \) was used for downstream comparisons against real motion \( \bm{D}_i^{(r)} \) and human-prompt-derived motion \( \bm{D}_i^{(b)} \).

\subsection{G-AI-HMS}
In this study, we used a T2M generation framework that integrates LLMs (GPT) with a dedicated motion synthesis architecture to improve motion realism and reduce kinematic errors. This combined approach is particularly advantageous for handling ambiguous or out-of-vocabulary phrases during prompt-based motion generation. The generative text model serves as a semantic guidance that helps the motion model's ability to generate physically plausible trajectories, even for novel or context-dependent actions. The full G-AI-HMS pipeline, including prompt enhancement, motion synthesis, pose extraction, normalization, and evaluation, is detailed in Algorithm 1.

\begin{algorithm}
\caption{G-AI-HMS: Generative AI-Driven High-Fidelity Human Motion Simulation}
\begin{algorithmic}[1]
\STATE \textbf{Input:} Task prompt set \( \mathcal{G} = \{g^(i)\}_{i=1}^N \), human video set \( \mathcal{D}^{(r)} = \{\bm{D}_i^{(r)}\}_{i=1}^N \)
\STATE \textbf{Output:} AI-generated motion sequences \( \{\bm{D}_i^{(s)}\}_{i=1}^N \), evaluation metrics \( \text{MPJPE}, \text{PA-MPJPE}, \text{DTW} \)

\STATE \textbf{Stage 1: Prompt Enhancement and Motion Synthesis}
\FOR{each prompt \( g^(i) \in \mathcal{G} \)}
    \STATE Enhance prompt using LLM: \( g_{\text{ai}}(i) \leftarrow \text{ChatGPT}(g^(i)) \)
    \STATE Generate motion from AI-enhanced prompt: \( \bm{D}_i^{(s)} \leftarrow \text{MotionGPT}(g_{\text{ai}}(i)) \)
    \STATE Generate motion from human prompt: \( \bm{D}_i^{(b)} \leftarrow \text{MotionGPT}(g^(i)) \)
\ENDFOR

\STATE \textbf{Stage 2: Human Motion Extraction}
\FOR{each human video \( \bm{D}_i^{(r)} \in \mathcal{D}_r \)}
    \STATE Extract joint trajectories using MediaPipe: \( X(i) \leftarrow \text{MediaPipe}(\bm{D}_i^{(r)}) \)
    \STATE Retain 22 common joints: \( \bm{D}_i^{(r)} \leftarrow \text{map\_to\_HumanML3D}(X(i)) \)
\ENDFOR

\STATE \textbf{Stage 3: Normalization and Filtering}
\FOR{each motion \( M \in \{\bm{D}_i^{(s)}, \bm{D}_i^{(b)}, \bm{D}_i^{(r)}\} \)}
    \STATE Root-center relative to head joint
    \STATE Normalize scale using distance between head and foot
    \STATE Flip \( y \)-axis for coordinate system consistency
    \STATE Apply median filter (kernel = 11)
    \STATE Apply Butterworth low-pass filter (cutoff = 0.05, order = 4)
\ENDFOR

\STATE \textbf{Stage 4: Temporal Alignment}
\STATE Determine minimum frame count: \( T^* \leftarrow \min(T_s, T_b, T_r) \)
\STATE Resample each motion \( M \in \{\bm{D}_i^{(s)}, \bm{D}_i^{(b)}, \bm{D}_i^{(r)}\} \) to length \( T^* \) via linear interpolation

\STATE \textbf{Stage 5: Evaluation}
\FOR{each motion source \( s \in \{\text{ai}, \text{human}\} \)}
    \STATE Compute MPJPE:
    \STATE \( \text{MPJPE}_s = \frac{1}{T^* \cdot 22} \sum_{t=1}^{T^*} \sum_{j=1}^{22} \| \mathbf{P}_{t,j}^{(s)} - \mathbf{P}_{t,j}^{(r)} \|_2 \)

    \STATE Compute PA-MPJPE:
    \STATE Align using Procrustes and compute \( \| R \cdot P_t^{\text{real}} - P_t^s \|_2 \)

    \STATE Compute DTW for each joint trajectory:
    \STATE \( \text{DTW}_s = \frac{1}{22} \sum_{j=1}^{22} \text{DTW}(P_{1:T^*}^s(j), P_{1:T^*}^{\text{real}}(j)) \)
\ENDFOR

\STATE \textbf{return} \( \{\bm{D}_i^{(s)}\}_{i=1}^N, \text{MPJPE}, \text{PA-MPJPE}, \text{DTW} \)

\end{algorithmic}
\end{algorithm}

The core model used for generating the AI-driven motion sequences evaluated in this work was MotionGPT \citep{ref11}, a motion-language model that uses a VQ-VAE as its motion tokenizer and a T5-based text-to-text transformer as its linguistic architecture. MotionGPT was not used as a final step in the analysis, but rather as the source for generating all AI-based motion data that were subsequently compared against MediaPipe reference sequences and human-generated prompts. Supporting its use in this study, Latreche et al. \citep{ref63} concluded that MediaPipe achieved excellent reliability across four shoulder joint movements, with intra-class correlation coefficients all more than 0.8 and mean joint angle differences from -0.01 degrees to almost -0.6 degrees when compared against goniometers and digital inclinometers. Similarly, Hamilton et al. \citep{ref64} showed that MediaPipe provided smooth and accurate tracking of joints, with coefficient of variation consistently below 10\% for most range-of-motion and joint angle time series measurements, with strong intra-class correlations with Qualisys 3D motion capture for flexion-extension tasks like elbow and knee movements.

The MotionGPT training pipeline consists of three stages. First, the VQ-VAE is trained to discretize motion sequences into latent motion tokens. Second, the text-to-text model is trained jointly on motion-language data, for learning cross-modal representations. The entire architecture is fine-tuned using natural language prompt instructions to enhance its alignment between linguistic semantics and spatiotemporal motion dynamics. This multi-stage process help MotionGPT generate diverse and realistic human motions from text descriptions by jointly modeling context, temporal continuity, and biomechanical plausibility \citep{ref11}.

\begin{figure}
\centering
\includegraphics[width=1\columnwidth]{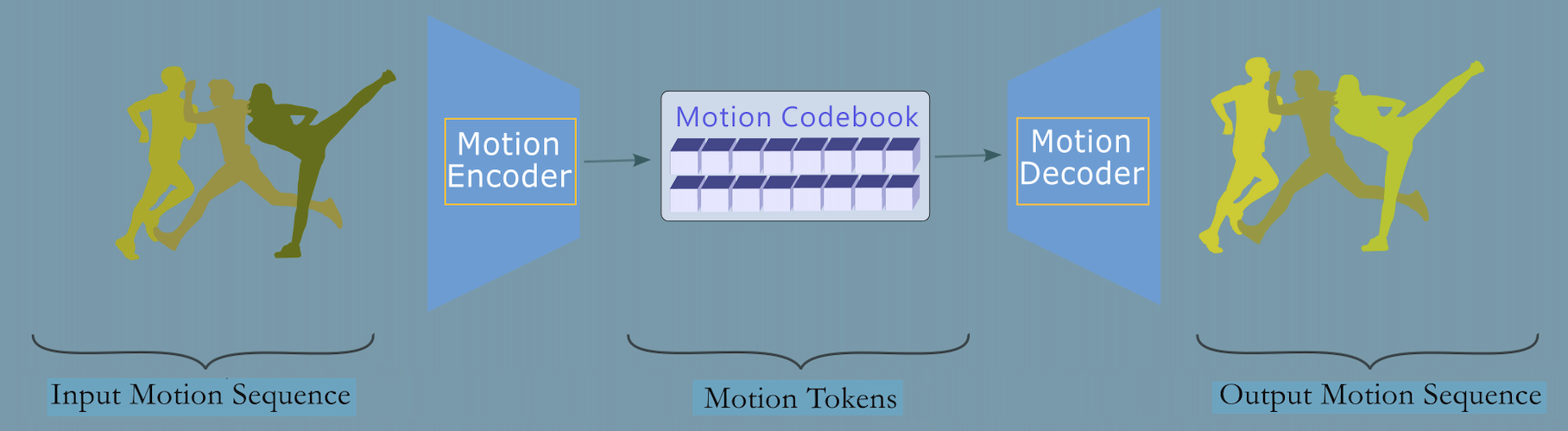}
\caption{Architecture of the motion tokenizer used in MotionGPT, based on the VQ-VAE framework \citep{ref11, ref44}.}
\label{fig:motion_tokenizer}
\end{figure}

\section{Data Analysis}

This section describes the full data processing pipeline implemented to compare motion sequences generated via three distinct modalities: AI-enhanced prompts (using ChatGPT), human-generated prompts, and reference motion data extracted from MediaPipe. The procedure involves motion synthesis (for AI and human prompt modalities), spatial normalization, temporal alignment, evaluation metric computation, and joint-level error analysis. Comparability across motion sequences was ensured by transforming them into a consistent format with aligned coordinate systems and frame counts.

\subsection{Spatial Normalization and Alignment}

\begin{figure}[H]
\centering
\includegraphics[width=0.7\linewidth]{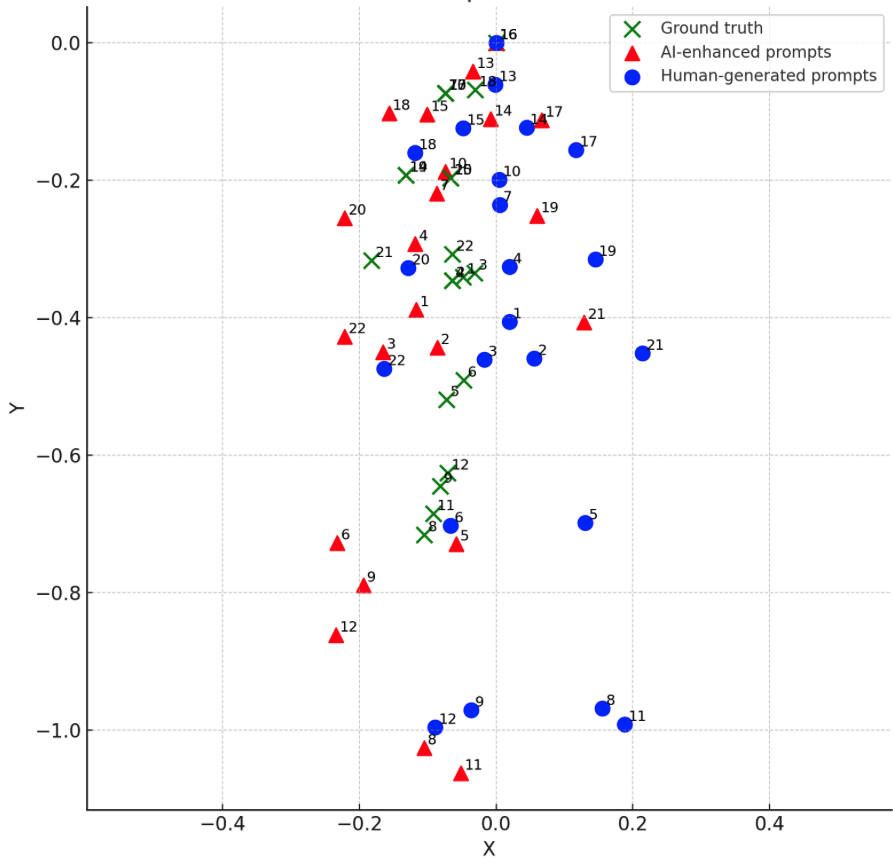}
\caption{Root-centered 2D poses with joint labels from a representative frame in Task 1, comparing AI-enhanced, human-generated, and MediaPipe ground truth motions.}
\end{figure}

Each motion sequence was modeled as a tensor, $\mathbf{M} \in \mathbb{R}^{T \times 22 \times 3}$, for $T$ frames, 22 anatomical joints, and 3 axes in Cartesian coordinates $(x, y, z)$. Figure 3 provides a visual comparison of the root-centered poses used in this representation.

\begin{table}[H]
\caption{Mapping of HumanML3D joint indices to corresponding MediaPipe landmark definitions used for alignment and evaluation.}
\begin{tabularx}{\textwidth}{|l|l|X|}
\hline
\textbf{ID} & \textbf{HumanML3D Joint} & \textbf{Corresponding MediaPipe Landmark} \\
\hline
1  & Root/Pelvis & MID\_HIP (average of LEFT\_HIP and RIGHT\_HIP) \\
\hline
2  & Right Hip & RIGHT\_HIP \\
\hline
3  & Left Hip & LEFT\_HIP \\
\hline
4  & Spine/Lumbar & Vector: MID\_HIP to MID\_SHOULDER \\
\hline
5  & Right Knee & RIGHT\_KNEE \\
\hline
6  & Left Knee & LEFT\_KNEE \\
\hline
7  & Spine/Thorax & MID\_SHOULDER \\
\hline
8  & Right Ankle & RIGHT\_ANKLE \\
\hline
9  & Left Ankle & LEFT\_ANKLE \\
\hline
10 & Spine/Upper & Midpoint between MID\_SHOULDER and NOSE \\
\hline
11 & Right Foot & RIGHT\_HEEL \\
\hline
12 & Left Foot & LEFT\_HEEL \\
\hline
13 & Neck Base & Vector from MID\_SHOULDER to NOSE \\
\hline
14 & Right Shoulder & RIGHT\_SHOULDER \\
\hline
15 & Left Shoulder & LEFT\_SHOULDER \\
\hline
16 & Head Top & NOSE \\
\hline
17 & Right Elbow & RIGHT\_ELBOW \\
\hline
18 & Left Elbow & LEFT\_ELBOW \\
\hline
19 & Right Wrist & RIGHT\_WRIST \\
\hline
20 & Left Wrist & LEFT\_WRIST \\
\hline
21 & Right Hand & RIGHT\_INDEX \\
\hline
22 & Left Hand & LEFT\_INDEX \\
\hline
\end{tabularx}
\end{table}

Raw video data was first processed using MediaPipe to extract 3D joint trajectories, represented as $\mathbf{X}_\text{ref} \in \mathbb{R}^{T \times 33 \times 3}$, where each frame contains 33 joint coordinates. To ensure compatibility with HumanML3D and AI-enhanced outputs, only the 22 common joints were retained, resulting in $\mathbf{X}_\text{aligned}^{\text{ref}} \in \mathbb{R}^{T \times 22 \times 3}$. The mapping between HumanML3D joints and their corresponding MediaPipe landmarks is summarized in Table 4.

The normalization involved two stages, root-centering and scale normalization. First, to spatially center the motion, each joint position $\mathbf{P}_{t,j}$ is translated relative to the head joint (joint 16) at each frame $t, t=1, ..., T$, producing $\tilde{\mathbf{P}}_{t,j}, j=1, ..., 22$, as shown in Equation (5).
\begin{equation}
\tilde{\mathbf{P}}_{t,j} = \mathbf{P}_{t,j} - \mathbf{P}_{t,16}
\end{equation}

Secondly, scale normalization was applied based on the Euclidean distance between the head (joint 16) and the left foot (joint 12), which served as a reference for body height. The scale factor $s_t$ for each frame $t$ is computed based on the Euclidean distance between the left foot (joint 12) and the head (joint 16), as shown in Equation (6). The normalized position of each joint is shown in Equation (7).
\begin{equation}
s_t = \| \mathbf{P}_{t,11} - \mathbf{P}_{t,16} \|_2
\end{equation}

\begin{equation}
\hat{\mathbf{P}}_{t,j} = \frac{\tilde{\mathbf{P}}_{t,j}}{s_t}
\end{equation}

Because MediaPipe adopts a top-left coordinate origin while other datasets use a bottom-left convention, the $y$-axis values in MediaPipe data were flipped for consistency. Equation (8) shows the flipping operation.
\begin{equation}
\hat{y}_{t,j} = -\hat{y}_{t,j}
\end{equation}

To further address high-frequency jitter and motion outliers introduced by sensor noise or generative drift, we applied an additional two-stage temporal filtering strategy. First, a median filter with kernel size $k = 11$ was used to remove impulsive spikes in each joint trajectory. A zero-phase fourth-order Butterworth low-pass filter with a normalized cutoff frequency of $f_c = 0.05$ was applied to smooth residual noise without introducing phase lag. For each joint $j$ and axis $d \in \{x, y, z\}$, the filtered signal $\tilde{P}_{t,j,d}^{\text{filtered}}$ was computed as shown in Equation (9).

\begin{align}
\tilde{P}_{t,j,d}^{\text{filtered}} =\ 
\text{Butter} \big( 
& \text{Median}\left( \hat{P}_{t,j,d},\ k{=}11 \right), \nonumber \\
& f_c{=}0.05,\ \text{order}{=}4 
\big)
\end{align}

This filtering step was applied to all motion sources $\hat{\mathbf{M}}^{(g)}$, $\hat{\mathbf{M}}^{(h)}$, and $\hat{\mathbf{M}}^{(m)}$ to improve biomechanical plausibility and ensure consistent signal smoothness across modalities.

These steps provided uniformly scaled and aligned motion tensors for all motion sources: AI-enhanced ($\hat{\mathbf{M}}^{(g)}$), human-generated ($\hat{\mathbf{M}}^{(h)}$), and ground truth video-based ($\hat{\mathbf{M}}^{(m)}$).

\subsection{Temporal Resampling}

Because different sources generated sequences of varying lengths, we applied linear interpolation to resample all motion sequences to the same number of frames. The target frame count $T^\ast$ was set to the minimum across the three modalities, as illustrated in Equation (10).  Here, $T_s$, $T_b$, and $T_r$ are the number of frames in the ground truth, human-observed, and model-generated motion sequences, respectively.
\begin{equation}
T^\ast = \min(T_s, T_b, T_r)
\end{equation}
For each coordinate sequence $x(t)$ over $T$ original frames, we defined an interpolation function. The function is shown in Equation (11). The resampled value at time step $t^\ast$ was calculated using Equation (12).
\begin{equation}
f(t) = \text{interp1d} \left( \frac{i}{T-1}, x_i \right), \quad i = 0, \ldots, T-1
\end{equation}

\begin{equation}
x^{\text{resampled}}(t^\ast) = f \left( \frac{t^\ast}{T^\ast - 1} \right), \quad t^\ast = 0, \ldots, T^\ast - 1
\end{equation}
where $\text{interp1d}$ is a function that performs single dimensional interpolation to construct a continuous mapping from discrete input-output pairs with linear interpolation. This generated temporally aligned sequences $\bar{\mathbf{M}}^{(g)}$, $\bar{\mathbf{M}}^{(h)}$, and $\bar{\mathbf{M}}^{(m)}$ for metric computation.

\subsection{Evaluation Metrics}

To assess the similarity between the generated and reference motions, we used three metrics: MPJPE, PA-MPJPE, and DTW.

MPJPE quantifies the average Euclidean distance between estimated and actual joint locations, aggregated across all joints and time frames, as defined in Equation (13).
\begin{equation}
\text{MPJPE} = \frac{1}{T^\ast \cdot 22} \sum_{t=1}^{T^\ast} \sum_{j=1}^{22} \left\| \bar{\mathbf{P}}_{t,j} - \bar{\mathbf{P}}^{(ground truth)}_{t,j} \right\|_2
\end{equation}
PA-MPJPE improves upon MPJPE by eliminating rigid-body transformations through the orthogonal Procrustes method. First, the joint matrices $\mathbf{A}$ and $\mathbf{B}$ were centered, as illustrated in Equation (14).
\begin{equation}
\tilde{\mathbf{A}} = \mathbf{A} - \bar{\mathbf{A}}, \quad \tilde{\mathbf{B}} = \mathbf{B} - \bar{\mathbf{B}}
\end{equation}
Then the optimal rotation matrix $\mathbf{R}$ was obtained via singular value decomposition. This operation is denoted in Equation (15).
\begin{equation}
\mathbf{H} = \tilde{\mathbf{B}}^\top \tilde{\mathbf{A}} = \mathbf{U} \Sigma \mathbf{V}^\top, \quad \mathbf{R} = \mathbf{U} \mathbf{V}^\top
\end{equation}
The aligned error was computed, as defined in Equation (16).
\begin{equation}
\text{PA-MPJPE} = \frac{1}{T^\ast \cdot 22} \sum_{t=1}^{T^\ast} \sum_{j=1}^{22} \left\| \bar{\mathbf{P}}_{t,j} - \mathbf{R} \cdot \bar{\mathbf{P}}^{(ground truth)}_{t,j} \right\|_2
\end{equation}

DTW measures the temporal alignment cost between two time-series. For each joint $j$, DTW was calculated between its 3D trajectories using Equation (17).
\begin{equation}
\text{DTW}_j = \text{DTW} \left( \{ \bar{\mathbf{P}}_{t,j} \}, \{ \bar{\mathbf{P}}^{(ground truth)}_{t,j} \} \right)
\end{equation}
The mean DTW score across joints was computed, as presented in Equation (18).
\begin{equation}
\text{DTW} = \frac{1}{22} \sum_{j=1}^{22} \text{DTW}_j
\end{equation}
As shown in the above equations, DTW captures both spatial and temporal misalignment.

\subsection{Joint-Level Error Metrics}

For fine-grained analysis, we also computed per-joint metrics. The per-joint MPJPE for joint $j$ is shown in Equation (19).
\begin{equation}
\text{MPJPE}_j = \frac{1}{T^\ast} \sum_{t=1}^{T^\ast} \left\| \bar{\mathbf{P}}_{t,j} - \bar{\mathbf{P}}^{(groundtruth)}_{t,j} \right\|_2
\end{equation}
The per-joint PA-MPJPE was calculated using Equation (20).
\begin{equation}
\text{PA-MPJPE}_j = \frac{1}{T^\ast} \sum_{t=1}^{T^\ast} \left\| \mathbf{R}_t \cdot \bar{\mathbf{P}}^{(groundtruth)}_{t,j} - \bar{\mathbf{P}}_{t,j} \right\|_2
\end{equation}
Per-joint DTW was then computed, as illustrated in Equation (21).
\begin{equation}
\text{DTW}_j = \text{DTW} \left( \{ \bar{\mathbf{P}}_{t,j} \}_{t=1}^{T^\ast}, \{ \bar{\mathbf{P}}^{(groundtruth)}_{t,j} \}_{t=1}^{T^\ast} \right)
\end{equation}

The complete analysis pipeline provided a framework for quantifying differences between motion generated by human-generated and AI-enhanced prompts relative to MediaPipe-based ground truth. The integration of MotionGPT and VQ-VAE was used for semantic-to-motion translation. The spatial and temporal alignment ensured accurate comparisons across modalities and tasks. This multi-stage approach supported detailed joint-level evaluations using well-established motion similarity metrics.

\section{Results}

The 3D human motions evaluated in this study were synthesized using MotionGPT, a generative model capable of translating natural language prompts into joint-based motion sequences. Both AI-enhanced and human-generated prompts were input to MotionGPT to produce corresponding motion outputs. These outputs were then quantitatively compared against ground truth motion references obtained using MediaPipe from video recordings of the eight target tasks.

\subsection{Task-Level Pose Similarity Metrics}

The comparison of AI-enhanced and human-generated prompts across eight tasks is summarized in Table 5. Metrics include MPJPE, PA-MPJPE, and DTW, where lower values indicate better alignment. For MPJPE, AI-enhanced prompts resulted in lower errors in six of the eight tasks: Task 1 (0.33 vs. 0.35), Task 4 (0.30 vs. 0.40), Task 5 (0.35 vs. 0.36), Task 6 (0.33 vs. 0.34), Task 7 (0.39 vs. 0.46), and Task 8 (0.35 vs. 0.36). Human-generated prompts showed better MPJPE in Tasks 2 and 3. In Task 4, the AI advantage was most pronounced, outperforming human prompts by over 0.09. For PA-MPJPE, human prompts produced lower errors in Tasks 2 (0.26 vs. 0.26), 3 (0.33 vs. 0.39), 4 (0.33 vs. 0.36), and 5 (0.29 vs. 0.29). AI-enhanced prompts were better in Tasks 1 (0.26 vs. 0.26), 6 (0.17 vs. 0.21), 7 (0.36 vs. 0.38), and values were nearly equivalent in Task 8 (0.35 vs. 0.35). Figures 4 and 5 illustrate the trajectories of ground truth, human-generated, and AI-enhanced prompts across all eight tasks.

In terms of temporal alignment, DTW scores were lower for AI-enhanced prompts in Tasks 1 (34.55 vs. 36.50), 3 (42.02 vs. 42.45), 4 (69.27 vs. 93.16), 5 (75.48 vs. 78.35), 6 (23.60 vs. 24.87), 7 (97.93 vs. 112.51), and 8 (96.24 vs. 98.43). The largest performance gap was again observed in Task 4, where AI prompts achieved a 23.89-point improvement over human prompts. Human-generated prompts had slightly lower DTW in Task 2 (49.58 vs. 51.99).

\begin{figure*}
    \centering
    \includegraphics[width=\textwidth, height=0.9\textheight, keepaspectratio]{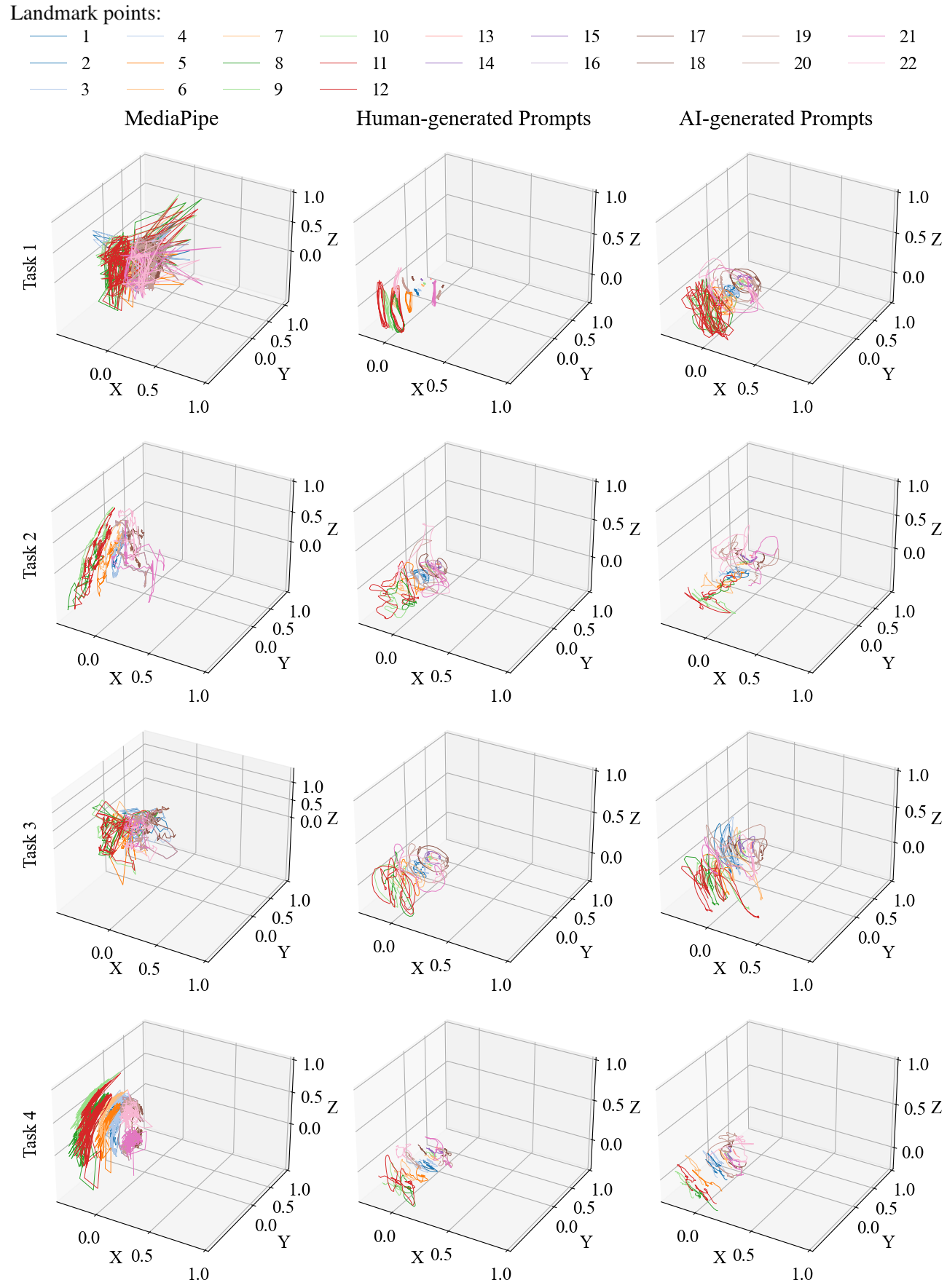}
    \caption{Trajectory comparison across Tasks 1 to 4: MediaPipe-generated ground truth motions, motions generated from human-written prompts, and motions generated from AI-enhanced prompts.}
\end{figure*}

\begin{figure*}
    \centering
    \includegraphics[width=\textwidth, height=0.9\textheight, keepaspectratio]{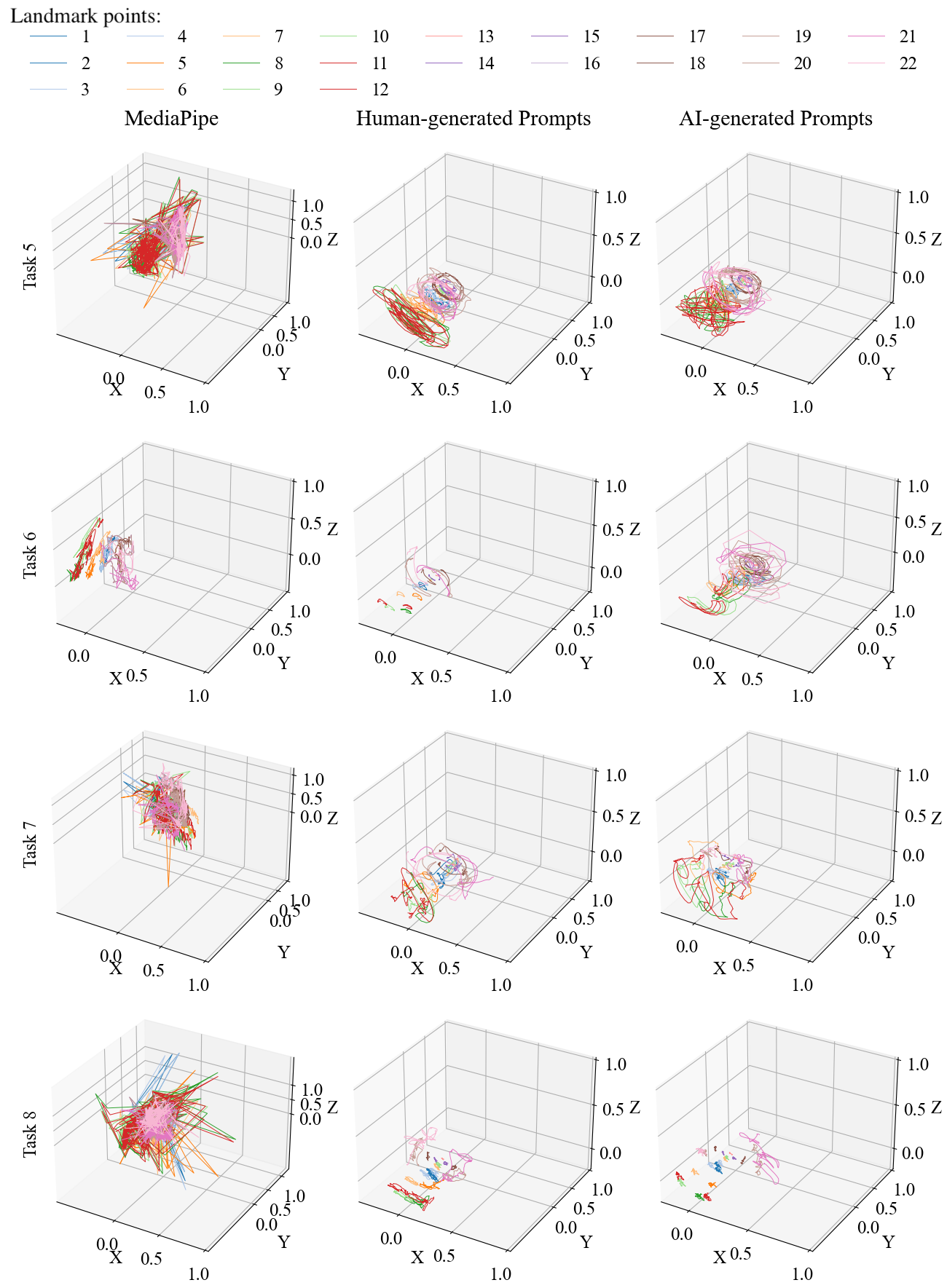}
    \caption{Trajectory comparison across Tasks 5 to 8: MediaPipe-generated ground truth motions, motions generated from human-written prompts, and motions generated from AI-enhanced prompts.}
\end{figure*}

\begin{table*}
\centering
\caption{Comparison of motion similarity metrics (MPJPE, PA-MPJPE, DTW) across eight tasks for AI-enhanced and human-generated prompts.}
\vspace{0.1cm}
\resizebox{\columnwidth}{!}{%
\begin{tabular}{|l|l|c|c|c|c|c|c|c|c|}
\hline
\textbf{Metric} & \textbf{Prompt} & \textbf{Task 1} & \textbf{Task 2} & \textbf{Task 3} & \textbf{Task 4} & \textbf{Task 5} & \textbf{Task 6} & \textbf{Task 7} & \textbf{Task 8} \\
\hline
\multirow{2}{*}{MPJPE} & AI-enhanced     & 0.33 & 0.37 & 0.39 & 0.30 & 0.35 & 0.33 & 0.39 & 0.35 \\ \cline{2-10}
                       & Human-generated & 0.35 & 0.34 & 0.38 & 0.40 & 0.36 & 0.34 & 0.46 & 0.36 \\
\hline
\multirow{2}{*}{PA-MPJPE} & AI-enhanced     & 0.26 & 0.26 & 0.39 & 0.36 & 0.29 & 0.17 & 0.36 & 0.35 \\ \cline{2-10}
                          & Human-generated & 0.26 & 0.26 & 0.33 & 0.33 & 0.29 & 0.21 & 0.38 & 0.35 \\
\hline
\multirow{2}{*}{DTW} & AI-enhanced     & 34.55 & 51.99 & 42.02 & 69.27 & 75.48 & 23.60 & 97.93 & 96.24 \\ \cline{2-10}
                     & Human-generated & 36.50 & 49.58 & 42.45 & 93.16 & 78.35 & 24.87 & 112.51 & 98.43 \\
\hline
\end{tabular}
}
\label{tab:overall_metrics}
\end{table*}

\subsection{Joint-Level Analysis by Task}

\begin{table*}
\centering
\caption{Joint-wise comparison of AI-enhanced and human-generated motions against MediaPipe ground truth averaged over tasks using MPJPE, PA-MPJPE, and DTW metrics.}
\vspace{0.1cm}
\resizebox{\columnwidth}{!}{%
\begin{tabular}{|c|l|c|c|c|c|c|c|}
\hline
\multirow{2}{*}{\textbf{ID}} & \multirow{2}{*}{\textbf{Landmark Name}} & 
\multicolumn{2}{c|}{\textbf{MPJPE}} & 
\multicolumn{2}{c|}{\textbf{PA-MPJPE}} & 
\multicolumn{2}{c|}{\textbf{DTW}} \\
\cline{3-8}
 & & \textbf{AI-enhanced} & \textbf{Human-generated} & \textbf{AI-enhanced} & \textbf{Human-generated} & \textbf{AI-enhanced} & \textbf{Human-generated} \\
\hline
1  & Root/Pelvis         & 0.29 & 0.30 & 0.12 & 0.12 & 50.31 & 54.75 \\ \hline
2  & Left Hip            & 0.31 & 0.32 & 0.15 & 0.15 & 52.46 & 56.75 \\ \hline
3  & Right Hip           & 0.34 & 0.35 & 0.20 & 0.20 & 58.03 & 63.01 \\ \hline
4  & Spine/Lumbar        & 0.32 & 0.34 & 0.21 & 0.20 & 56.56 & 61.03 \\ \hline
5  & Left Knee           & 0.37 & 0.38 & 0.26 & 0.25 & 65.61 & 69.13 \\ \hline
6  & Right Knee          & 0.41 & 0.44 & 0.32 & 0.29 & 66.39 & 76.87 \\ \hline
7  & Spine/Thorax        & 0.29 & 0.29 & 0.28 & 0.25 & 52.97 & 54.32 \\ \hline
8  & Left Ankle          & 0.49 & 0.51 & 0.36 & 0.36 & 86.24 & 93.95 \\ \hline
9  & Right Ankle         & 0.61 & 0.62 & 0.45 & 0.41 & 94.21 & 103.69 \\ \hline
10 & Spine/Upper         & 0.27 & 0.27 & 0.27 & 0.24 & 48.98 & 50.02 \\ \hline
11 & Left Foot           & 0.49 & 0.49 & 0.36 & 0.33 & 87.23 & 90.69 \\ \hline
12 & Right Foot          & 0.58 & 0.58 & 0.44 & 0.40 & 91.00 & 99.61 \\ \hline
13 & Neck Base           & 0.23 & 0.23 & 0.27 & 0.24 & 37.80 & 40.22 \\ \hline
14 & Left Shoulder       & 0.33 & 0.34 & 0.35 & 0.34 & 57.69 & 61.72 \\ \hline
15 & Right Shoulder      & 0.37 & 0.39 & 0.35 & 0.38 & 65.05 & 68.82 \\ \hline
16 & Head Top            & 0.00 & 0.00 & 0.25 & 0.25 & 0.00  & 0.00  \\ \hline
17 & Left Elbow          & 0.27 & 0.30 & 0.28 & 0.29 & 46.77 & 53.90 \\ \hline
18 & Right Elbow         & 0.27 & 0.31 & 0.30 & 0.31 & 45.51 & 54.00 \\ \hline
19 & Left Wrist          & 0.38 & 0.40 & 0.36 & 0.37 & 71.13 & 72.43 \\ \hline
20 & Right Wrist         & 0.37 & 0.44 & 0.36 & 0.39 & 67.10 & 79.72 \\ \hline
21 & Left Hand           & 0.41 & 0.45 & 0.41 & 0.44 & 77.71 & 83.37 \\ \hline
22 & Right Hand          & 0.38 & 0.43 & 0.39 & 0.44 & 71.72 & 85.56 \\ \hline
\end{tabular}
}
\end{table*}

Joint-wise comparisons showed detailed patterns across tasks. In Task 1 (walking back and forth), AI-enhanced prompts produced lower MPJPE in 20 of 22 joints, including both lower body joints such as the left ankle (0.50 vs. 0.57) and right ankle (0.50 vs. 0.57), as well as upper body joints like the left elbow (0.27 vs. 0.29). PA-MPJPE and DTW were also consistently lower for AI across joints, especially for joints involved in gait such as knees and feet. Figures 6 and 7 show the joint-wise similarity metrics (MPJPE, PA-MPJPE, and DTW) for ground truth, human-generated, and AI-enhanced motions across all eight tasks. Table 6 presents a joint-wise comparison of AI-enhanced and human-generated motions against MediaPipe ground truth for Task 8, using MPJPE, PA-MPJPE, and DTW metrics.

In Task 2 (throwing a large object), human-generated prompts showed lower MPJPE in 14 of 22 joints, particularly the right wrist (0.88 vs. 0.65) and right hand (0.93 vs. 0.70). PA-MPJPE and DTW followed a similar trend, suggesting more accurate representations of hand-dominant actions in human prompts. However, AI prompts performed comparably at the root and shoulder joints. For Task 3 (walking and sitting), human-generated prompts demonstrated better performance in PA-MPJPE and DTW across nearly all joints, including significant improvements in the hip, knee, and wrist regions. MPJPE differences were marginal, with AI performing slightly better on the pelvis and spine joints but worse on distal joints. In Task 4 (horizontal painting), AI-enhanced prompts substantially outperformed human prompts across almost all joints and metrics. For example, the MPJPE for the left elbow was 0.24 (AI) vs. 0.32 (human), and the right wrist DTW was 95.01 (AI) vs. 131.42 (human). These differences were consistent across upper limb joints, indicating greater fidelity of repetitive upper body actions under AI guidance. Task 5 (walking with a box on the head) showed mixed results. AI prompts had lower MPJPE for central joints like the pelvis and knees but performed worse in the foot and hand regions. Human prompts achieved lower PA-MPJPE in wrist and hand joints, which may reflect more accurate modeling of compensatory hand movement for balance. In Task 6 (throwing a small object), both AI and human prompts produced low joint-wise errors, with AI slightly outperforming in root and elbow joints but underperforming in right-hand joints. Task 7 (walking and sitting on a swivel chair) was characterized by relatively high DTW scores for both prompt types across joints. AI prompts had slightly lower MPJPE in the pelvis and hips, but human prompts showed lower DTW in hands and shoulders. This suggests the seated motion and rotation element introduced increased kinematic complexity for both modalities. Task 8 (vertical painting) showed a general trend of higher error magnitudes across metrics for both prompt types, likely due to the challenging nature of modeling vertical reach and brush motion. AI prompts showed better MPJPE and DTW for root, hip, and lower extremity joints, while human prompts were marginally better for upper extremities including the right wrist and hand. Both methods struggled with high DTW and PA-MPJPE in the ankle and foot joints, with values exceeding 170 in many cases.

A statistical comparison between AI-enhanced and human-generated prompts across three joint-wise metrics, MPJPE, PA-MPJPE, and DTW, revealed significant differences in two of the three measures. Paired t-tests indicated that AI-enhanced prompts resulted in significantly lower MPJPE (0.353 vs. 0.375; t = –4.20, p < 0.0001) and DTW scores (61.39 vs. 67.01; t = –5.32, p < 0.0001), suggesting better spatial and temporal alignment with the ground truth. In contrast, no significant difference was observed for PA-MPJPE (t = 0.64), indicating comparable performance in pose-aligned joint error. Normality of the paired differences was verified using Shapiro–Wilk tests.

\begin{figure*}
    \centering
    \includegraphics[width=\textwidth, height=0.9\textheight, keepaspectratio]{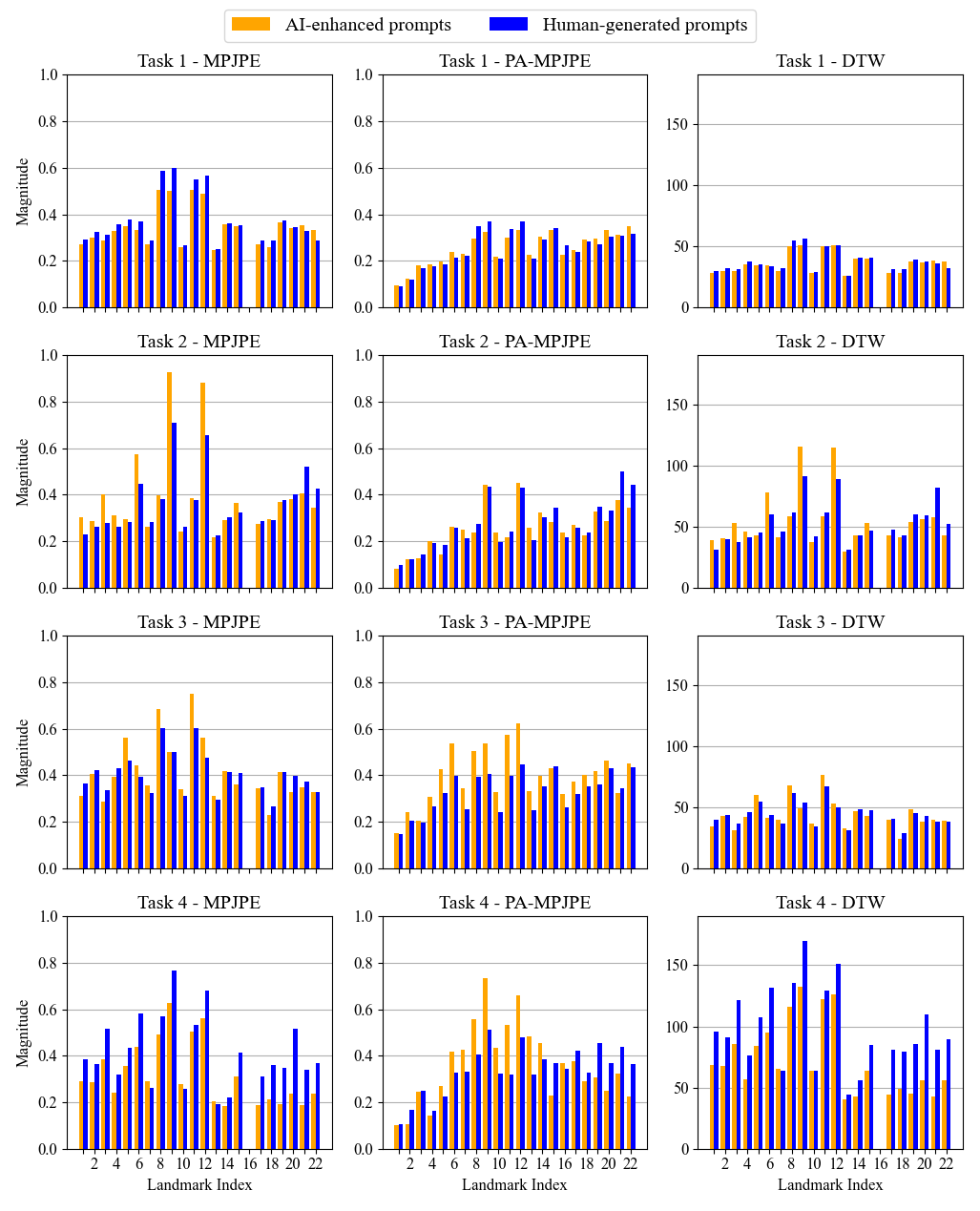}
    \caption{Landmark point-wise comparison of motion similarity metrics (MPJPE, PA-MPJPE, DTW) for Tasks 1–4. Each subplot shows metric values for 22 anatomical landmarks, comparing AI-enhanced prompts against human-generated prompts.}
\end{figure*}

\begin{figure*}
    \centering
    \includegraphics[width=\textwidth, height=0.9\textheight, keepaspectratio]{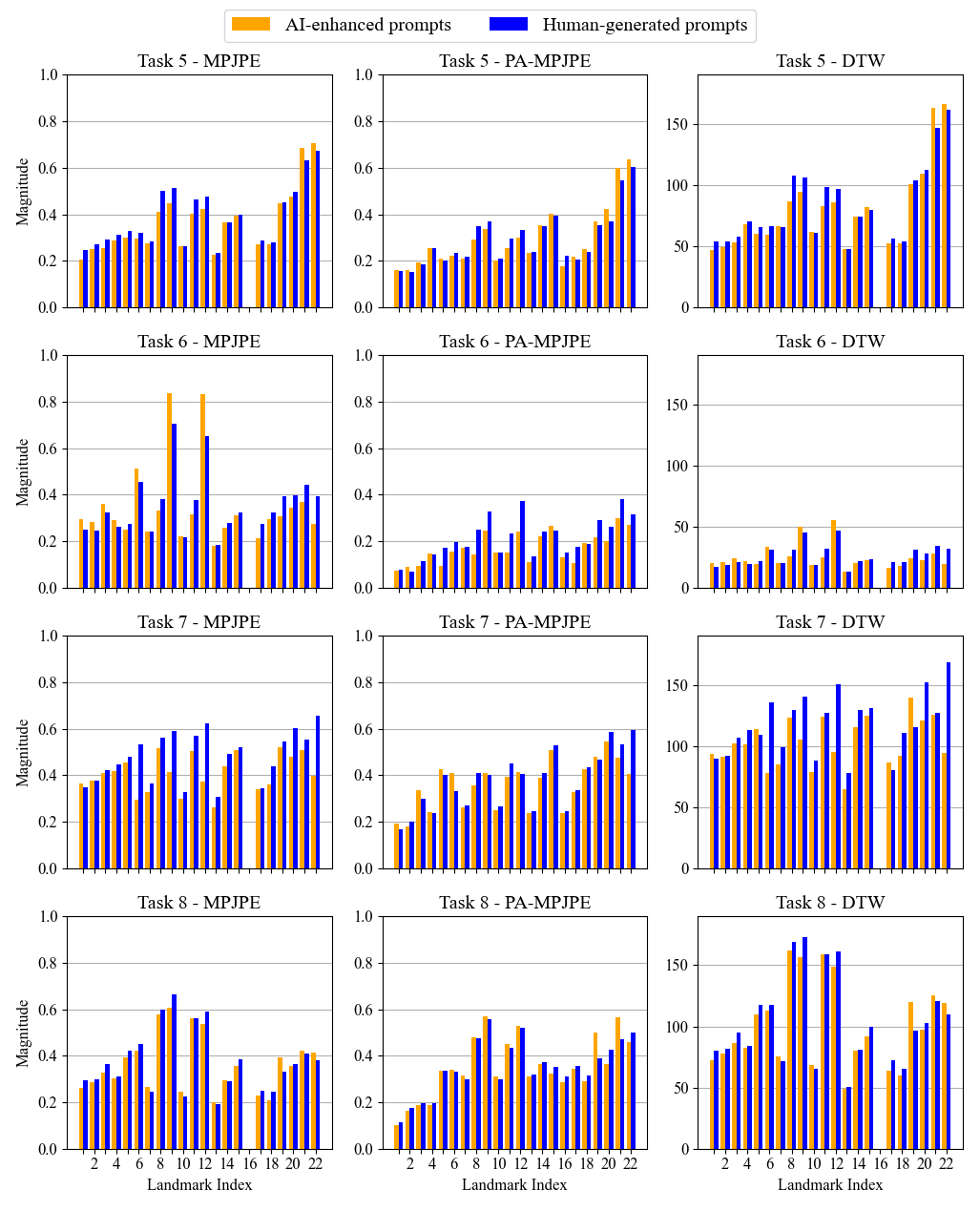}
    \caption{Landmark point-wise comparison of motion similarity metrics (MPJPE, PA-MPJPE, DTW) for Tasks 5-8. Each subplot shows metric values for 22 anatomical landmarks, comparing AI-enhanced prompts against human-generated prompts.}
\end{figure*}

\section{Discussion}

\subsection{Overview of Performance Trends}

This study evaluated the comparative effectiveness of AI-enhanced and human-generated natural language prompts for motion generation across eight diverse physical tasks using three widely adopted metrics: MPJPE, PA-MPJPE, and DTW. All motion sequences were generated using MotionGPT, a transformer-based model that converts natural language descriptions into 3D joint trajectories. Both prompt types were input to MotionGPT under identical settings to ensure fair comparison. The synthesized motions were then evaluated against ground truth references obtained using MediaPipe from corresponding task videos. The quantitative results provide evidence that AI-enhanced prompts can generate better kinematic outputs compared to human prompts, particularly in tasks with cyclic, gross motor characteristics. AI-enhanced prompts achieved lower MPJPE in 6 of 8 tasks, lower PA-MPJPE in 4 tasks, and lower DTW in 7 tasks, highlighting their strength in capturing both spatial and temporal structure. However, performance varied considerably by task type and body region, showing distinct strengths and limitations of each prompt source \citep{ref1, ref2}.

\begin{table}
\centering
\caption{Statistical comparison between AI-enhanced and human-written prompts across all joint-wise metrics. Paired t-tests were used for significance testing.}
\vspace{0.1cm}
\begin{tabular}{|l|c|c|c|}
\hline
\textbf{Metric} & \multicolumn{2}{c|}{\textbf{Mean}} & \textbf{t} \\
\cline{2-3}
                & \textbf{AI-enhanced} & \textbf{Human-generated} & \\
\hline
MPJPE     & 0.353   & 0.375   & -4.20$^{*}$ \\
\hline
PA-MPJPE  & 0.306   & 0.303   &  0.64 \\
\hline
DTW       & 61.39   & 67.01   & -5.32$^{*}$ \\
\hline
\end{tabular}
\vskip 0.5em  
\begin{tablenotes}
\item $^{*}$Significant at $p < 0.0001$ (Shapiro–Wilk tests confirmed normality of paired differences).
\end{tablenotes}
\end{table}

As shown in Table 7, paired comparisons across joint-wise metrics revealed that AI-enhanced prompts produced significantly lower MPJPE and DTW values ($p < 0.0001$), demonstrating better spatial accuracy and temporal consistency. The lack of a significant difference in PA-MPJPE ($p = 0.52$) indicates that both prompt types yielded comparable pose structures. These results suggest that AI-generated prompts enhance motion fidelity without distorting joint alignment.

\subsection{Scientific Basis and Implications of Evaluation Metrics}

Each evaluation metric describes a different dimension of motion accuracy. MPJPE quantifies the Euclidean distance between predicted and reference joint positions in 3D space, making it a strict spatial fidelity measure. PA-MPJPE removes rigid-body transformations (scale, rotation, and translation), focusing on pose structure and relative limb configuration \citep{ref1}. Lower MPJPE indicates accurate global positioning, whereas lower PA-MPJPE reflects structurally coherent body postures. The divergence observed between MPJPE and PA-MPJPE in several tasks (e.g., Task 3 and 4) highlights that spatial accuracy and postural fidelity are not always aligned, particularly when body orientation or task-specific postures are not explicitly encoded in the prompt \citep{ref11, ref60}. DTW evaluates temporal alignment, capturing how well joint trajectories evolve over time. It is particularly sensitive to phase errors, misaligned motion cycles, and unnatural transitions \citep{ref50}. Lower DTW scores suggest smoother, time-consistent motion. These three metrics provide a framework for evaluating both kinematic realism and biomechanical structure. The fact that AI-enhanced prompts achieved strong MPJPE and DTW performance in certain tasks but lagged in PA-MPJPE in others suggests that while global motion is often well-conveyed, articulation at the limb level, particularly in fine motor tasks, remains an area of challenge \citep{ref1, ref3}.

\subsection{Task Complexity and Prompt Efficacy}

Performance varied not only by metric but also by task type. In locomotor tasks such as walking (Task 1) and walking-while-carrying (Task 5), AI prompts consistently showed lower MPJPE and DTW, suggesting reliable capture of rhythmic and symmetrical body movements. These findings are consistent with previous research on motion generation where cyclic tasks, due to their lower entropy and repetitive structure, are easier to model via autoregressive or transformer-based architectures \citep{ref11, ref29}. Joint-wise results reinforced this trend: pelvis, hip, and ankle trajectories were often more accurately synthesized under AI control. In tasks involving tool use, asymmetry, or multiple action phases, such as painting a vertical wall (Task 8) or sitting in a rotating chair (Task 7), AI-enhanced prompts resulted in greater variance and mixed or higher PA-MPJPE, despite showing lower DTW in both tasks. These tasks inherently involve complex transitions (e.g., initiation of seated descent), anticipatory stabilization (e.g., reaching), and subtle wrist and spine coordination, which are biomechanically detailed behaviors that human writers, informed by lived experience, may implicitly encode in natural language \citep{ref3, ref4, ref41}. Human prompts provided more accurate articulation of distal joints like wrists, hands, and elbows in these contexts. Task 4 (painting a flat surface) was a hybrid case. Although the upper limbs were important, the task’s repeated flat movements and small vertical changes made it easier for AI prompts to create accurate motion paths. Joint-wise comparisons confirmed superior performance for AI prompts across shoulder, elbow, and wrist MPJPE, PA-MPJPE, and DTW in this task. This suggests that even in upper-limb tasks, prompt interpretability hinges on the predictability and cyclicity of joint movement rather than body part alone \citep{ref26}.

\subsection{Anatomical Fidelity and Joint-Wise Dynamics}

A fine-grained analysis joint-wise performance shows that AI-enhanced prompts have better fidelity for proximal joints and trunk alignment. Across most tasks, lower MPJPE and DTW were observed for pelvis, spine, and thorax regions. This may result from the general structure-preserving tendencies of pretrained language models and their tendency to default to "average" or biomechanically typical postures. Proximal joint control benefits from the relative regularity of base-of-support motion patterns in LLM training data, which includes walking, standing, and gesturing \citep{ref1, ref10}. However, distal joints (wrists, feet, and fingers) posed consistent challenges. The wrist and ankle often had the largest discrepancies across tasks, especially in orientation-sensitive metrics such as PA-MPJPE and DTW. For example, in Tasks 2 and 6, AI-enhanced wrist motions displayed temporal inconsistencies and misaligned trajectories, resulting in elevated PA-MPJPE and DTW scores. This reflects known issues in generative motion synthesis where distal limb trajectories are harder to anchor without explicit context such as object affordance, environmental interaction, or inverse kinematics constraints \citep{ref5, ref22, ref27}. The high error variance in head and neck alignment (e.g., Joint 16) despite having low MPJPE (often 0) is indicative of static reference points being preserved in position but failing in pose estimation. This points to a limitation in prompt guidance when orientation is not specified \citep{ref11, ref12}.

\subsection{Metric Cross-Analysis and Complementarity}

The differential behavior of metrics under various prompts emphasizes the need for multi-metric evaluations. For example, AI-enhanced prompts often achieved lower MPJPE but higher PA-MPJPE, indicating that although joint locations were close to the target positions, the relative body posture was sometimes misaligned. Human prompts that showed higher MPJPE but lower PA-MPJPE, as in Task 2, may reflect natural joint positioning but offset body translations \citep{ref1, ref3}. Similarly, DTW often diverged, with AI-enhanced prompts achieving lower scores in 7 of 8 tasks, even when spatial errors were present—highlighting their strength in modeling temporal structure.

\subsection{Case Study I: Practical Use of Low-Error Simulations}

Tasks with low MPJPE and DTW scores, such as walking (Task 1) or horizontal painting (Task 4), have strong potential for downstream applications in occupational training, animation, or digital ergonomics. In such contexts, precise orientation is less critical than consistent, repeatable motion patterns. AI-enhanced simulations, with syntactically well-structured and semantically rich descriptions, were capable of producing these motion sequences with sufficient fidelity \citep{ref3, ref4, ref20}. For example, the simulated action of “lifting a box next to your right leg and placing it next to your left leg” generated coherent pelvic and hand trajectories even in the absence of explicit object modeling \citep{ref14, ref28}. These capabilities are highly relevant for applications in virtual worker training, animated procedural tutorials, or digital twins in manufacturing environments where scene reconstruction from language may supplement or replace manual animation \citep{ref9, ref12, ref41, ref57}.

\subsection{Case Study II: Instruction Tuning for Prompt-to-Motion Alignment}

Instruction tuning (IT) is a method for aligning LLM-generated prompts more closely with downstream motion synthesis objectives. IT involves fine tuning a pretrained model on a dataset of instructional pairs, comprising text prompts and corresponding responses, to improve its ability to follow tasks and adapt to specific domains \citep{ref46}. In this study, the AI-enhanced prompts were created using a GPT-style LLM that was instruction-tuned to maximize semantic completeness and biomechanical clarity \citep{ref11, ref10}. Instruction-tuned prompts resulted in improved structural regularity and descriptive sufficiency, especially in tasks with cyclic trajectories and full-body symmetry, such as walking or box-lifting. However, limitations remain in representing inter-limb dependencies or non-linear transitions, as reflected in higher PA-MPJPE in upper-limb-dominant tasks like painting vertically or sitting with rotation. This suggests that future iterations of instruction tuning for motion applications should incorporate multimodal finetuning with both language and motion feedback (e.g., velocity constraints, biomechanical realism losses), as has been explored in robot instruction following \citep{ref21, ref30}. IT also makes it possible to fluently adjust and scale different motion styles. In practice, LLMs can be trained to generate prompts tailored to specific personas, such as an "ergonomics trainer" or "motion capture director," to emphasize attributes like safety, clarity, or expressive range. Such fine control over prompt structure may mitigate current inconsistencies in distal limb modeling and provide domain-specific prompt templates that generalize across task families \citep{ref9, ref46}.

\subsection{Toward Hybrid Prompting Systems}

The divergence in strengths between AI and human-generated prompts supports the development of hybrid prompting systems. In this approach, AI-enhanced prompts can serve as a base layer, ensuring linguistic consistency and strong temporal coherence, while domain experts refine the descriptions to address limb-specific articulation, fine motor control, or spatial orientation. This approach is similar to co-creative systems in animation and design, where humans and AI work together by refining generated outputs step by step \citep{ref20, ref25}. In critical applications such as construction training, rehabilitation, or collaborative robotics, where improper articulation or temporal delays may pose risks, human oversight of generated motions remains essential. However, for scalable content creation in digital avatars, educational games, or ergonomic simulations, instruction-tuned AI prompts provide a strong starting point with good generalization for repetitive, full-body tasks and minimal manual effort \citep{ref8, ref10, ref41}.

\subsection{Study Limitations}

While this study demonstrates the feasibility of using LLM prompts for accurate motion generation, several limitations should be acknowledged. First, MediaPipe was used as the motion reference, but its limitations in pose estimation accuracy, particularly under occlusion, rapid movement, or depth variation, may affect the reliability of the ground truth \citep{ref48}. Second, the study assumed one-to-one mapping between a static prompt and a fixed reference motion; future work should explore multi-prompt aggregation, dynamic prompt modulation, and feedback-conditioned synthesis. Third, the evaluation inherently compares AI-enhanced motions to human-generated ground truth, introducing an unavoidable bias given the subjective characteristics of human movement. Future work could address this limitation by developing a more comprehensive video dataset that includes motions generated both from human-designed prompts and AI-enhanced prompts, for a more balanced and systematic comparison. Expanding the prompt vocabulary to include task modifiers (e.g., “quickly,” “cautiously,” “with shoulder lead”) or environmental context may provide finer control of motion variability, which is essential for human-centered applications like exoskeleton testing or industrial training simulation \citep{ref41, ref20}. Integration with simulation environments or physics-based constraints would further enhance realism and applicability \citep{ref31, ref32}.

\section{Conclusion and Future Work}

This study compared AI-enhanced and human-generated natural language prompts for 3D human motion synthesis across eight tasks using three metrics: MPJPE, PA-MPJPE, and DTW. Results show that AI-enhanced prompts, can produce motion outputs with comparable or better accuracy to human-written descriptions in tasks involving gross motor patterns and repetitive actions. However, human prompts showed advantages in fine-grained articulation of distal joints and complex transitions, particularly in tasks involving asymmetric movement or tool use. Joint-wise analyses showed that AI-enhanced prompts tend to preserve spatial regularity in proximal joints, while greater variability is observed in the wrist, ankle, and hand regions, likely due to limited contextual detail in the prompt descriptions. Differences between metrics highlight the importance of using multiple evaluation methods to assess motion realism and consistency.

Future work should explore multimodal fine-tuning strategies incorporating biomechanical feedback, contextual object cues, and dynamic prompt structuring. Integrating AI-enhanced prompts with human-in-the-loop editing, style conditioning, and domain-specific lexical templates may improve reliability in safety-critical applications. Expanding task diversity and benchmarking against higher-fidelity motion capture systems could improve generalizability and support broader adoption in training, simulation, and interactive environments. The G-AI-HMS codebase, adapted from MotionGPT \citep{ref11}, is available on GitHub \citep{ref52}.

\section*{Acknowledgment}

The authors would like to thank the participant who took the time to perform the experiments that were a part of the current study.

\section*{CRediT Author Statement}

\noindent Hari Iyer: Conceptualization, Methodology, Software, Writing – original draft preparation, Visualization.\\
Neel Macwan: Data curation, Formal analysis, Validation, Writing – review and editing.\\
Atharva Jitendra Hude: Investigation, Software, Validation, Resources.\\
Heejin Jeong: Project administration, Methodology, Writing – review and editing.\\
Shenghan Guo: Supervision, Software, Formal analysis, Writing – review and editing.

\end{document}